\documentclass[journal]{IEEEtran}
\usepackage{cite}

\ifCLASSINFOpdf
\else
   \usepackage[pdflatex]{graphicx}
\fi
\usepackage{amsmath,amssymb,supertabular}
\usepackage[pdftex]{graphicx}
\newcommand{\bm}[1]{\mbox{\boldmath $#1$}}

\begin{document}
%
\title{A Recurrent Probabilistic Neural Network with Dimensionality Reduction Based on Time-series Discriminant Component Analysis}
%
%
%

\author{Hideaki~Hayashi,~\IEEEmembership{Member,~IEEE,}
		Taro~Shibanoki,~\IEEEmembership{Member,~IEEE,}
        Keisuke~Shima,~\IEEEmembership{Member,~IEEE,}
        Yuichi~Kurita,~\IEEEmembership{Member,~IEEE,}
        and~Toshio~Tsuji,~\IEEEmembership{Member,~IEEE}
\thanks{H. Hayashi is with Department of Advanced Information Technology, Kyushu University, 744, Motooka, Nishi-ku, Fukuoka-shi, 819-0395 JAPAN.
        e-mail: hayashi@ait.kyushu-u.ac.jp}
\thanks{T. Shibanoki is with College of Engineering, Ibaraki University, Hitachi, Japan.}
\thanks{K. Shima is with Faculty of Engineering, Yokohama National University, Yokohama, 240-8501 Japan.}
\thanks{Y. Kurita and T. Tsuji 
		are with Institute of Engineering, Hiroshima University, Higashi-hiroshima, 739-8527 Japan}
\thanks{ }}

%
%

\markboth{ }%
{Shell \MakeLowercase{\textit{et al.}}: Bare Demo of IEEEtran.cls for Journals}
%



\maketitle

\begin{abstract}
This paper proposes a probabilistic neural network developed on the basis of time-series discriminant component analysis (TSDCA) 
that can be used to classify high-dimensional time-series patterns. TSDCA involves the compression of high-dimensional time series 
into a lower-dimensional space using a set of orthogonal transformations and the calculation of posterior probabilities based on 
a continuous-density hidden Markov model with a Gaussian mixture model expressed in the reduced-dimensional space. 
The analysis can be incorporated into a neural network, which is named a time-series discriminant component network (TSDCN), 
so that parameters of dimensionality reduction and classification can be obtained simultaneously as network coefficients 
according to a backpropagation through time-based learning algorithm with the Lagrange multiplier method. 
The TSDCN is considered to enable high-accuracy classification
of high-dimensional time-series patterns and to reduce the computation time taken for network training. 
The validity of the TSDCN is demonstrated for high-dimensional artificial data 
and EEG signals in the experiments conducted during the study.
\end{abstract}


%
\IEEEpeerreviewmaketitle

{\allowdisplaybreaks
\section*{Nomenclature}
\addcontentsline{toc}{section}{Nomenclature}
\begin{IEEEdescription}[\IEEEusemathlabelsep\IEEEsetlabelwidth{$V_1,V_2,V_3$}]
\item[$D$]										Dimensionality in the original space
\item[$D'$]										Dimensionality in the subspace 
\item[$C$]										Number of classes 
\item[$K_c$]									Number of states 
\item[$M_{c,k}$]								Number of components 
\item[$P(\cdot)$]								Probability	
\item[$\bm{x}(t)$]								Time-series vector in the original space 
\item[$\bm{x'}(t)$]								Time-series vector in the subspace 
\item[${\bf V}^{(c,k,m)}$]						Orthogonal transformation matrix 
\item[$v^{(c,k,m)}_{i,j}$]						Element of ${\bf V}^{(c,k,m)}$ 
\item[$\bm{\mu}^{(c,k,m)}$]						Mean vector in the original space 
\item[$\bm{\mu'}^{(c,k,m)}$]					Mean vector in the subspace 
\item[$g(\cdot)$]								Gaussian distribution 
\item[$r_{c,k,m}$]								Mixture proportion of a GMM 
\item[${\Sigma'}^{(c,k,m)}$]					Covariance matrix in the subspace 
\item[$s'^{(c,k,m)}_{i,j}$]						Element of $({\Sigma'}^{(c,k,m)})^{-1}$ 
\item[$\gamma^c_{k',k}$]						State change probability of an HMM 
\item[$\pi_k^c$]								Prior probability of an HMM 
\item[${\bf X}(t)$]								Transformed vector in the original space 
\item[${\bf X'}^{(c,k,m)}(t)$]					Transformed vector in the subspace 
\item[${\bf W}^{(c,k,m)}$]						Weight between the first/second layers 
\item[$w_{i,j}^{(c,k,m)}$]						Element of ${\bf W}^{(c,k,m)}$ 
\item[${\bf W'}^{(c,k',k,m)}$]					Weight between the third/fourth layers 
\item[${w'}_h^{(c,k',k,m)}$]					Element of ${\bf W'}^{(c,k',k,m)}$ 
\item[${}^{(i)}I$]								Input of the $i$th unit 
\item[${}^{(i)}O$]								Output of the $i$th unit 
\item[${\bm Q}^{(n)}$]							Teacher vector 
\item[$J$]										Negative log-likelihood function 
\item[$L$]										Lagrange function 
\item[$\bm{\lambda}^{(c,k,m)}$]					Lagrange multiplier 
\item[$\bm{h}^{(c,k,m)}$]						Orthogonality conditions 
\item[$\bm{d}^{(c,k,m)}_l$]						Modification amount for ${\bf W}^{(c,k,m)}$ 
\end{IEEEdescription}

\section{Introduction}
\IEEEPARstart{T}{ime-series} pattern classification has a wide range of applications 
such as speech recognition, gesture recognition, and biosignal classification, 
and many studies have been performed to investigate higher classification performance
\cite{radovanovic2010time,spiegel2011pattern,jeong2011weighted,buza2011insight,Prekopcsak2012,lines2012shapelet,xu2008video}.

Time-series pattern classification methods can be divided into three large categories --- 
sequence distance-based classification, feature-based classification, and model-based classification\cite{xing2010brief}. 
Sequence distance-based methods measure the similarity between a pair of patterns based on a distance function 
such as the Euclidean distance, the Mahalanobis distance\cite{xiang2008learning,xing2002distance}, or dynamic time warping, 
and then classify the patterns using conventional classification algorithms typified by a $k$-nearest neighbor classifier. 
In feature-based methods, features are extracted from the original time series and are classified using a support vector machine, 
decision trees, and neural networks (NNs). 
In particular, NNs are expanded for time-series classification, known as the appearance of the Jordan network\cite{jordan1997serial}, 
the Elman network \cite{elman1990finding}, and time delay neural networks \cite{waibel1989phoneme}. 
The most popular approach in model-based methods is the hidden Markov model (HMM).
In the HMM, the system for each class is modeled by a Markov process with unobserved states. 
Time-series patterns are then classified into the classes based on a likelihood that is calculated from the model. 
All the above methods, however, have some drawbacks: 
Sequence distance-based methods and model based methods need a large amount of training data 
to estimate the distribution of input data precisely. 
Feature-based methods are likely to cause overfitting because they have too many free parameters and complex structures.

In recent years, a fourth option, ``model-based NNs'', has been proposed as a hybrid of NNs and model-based methods 
\cite{caelli1993model,caelli1999modularity,fontaine2001model,perlovsky1991maximum,specht1991general,streit1994maximum,tsuji2003recurrent}. 
Model-based NNs are developed by integrating prior knowledge of the input data into the network structure 
to be capable of saving the amount of training data and preventing overfitting. 
Tsuji {\it et al.} \cite{tsuji2003recurrent} proposed a recurrent probabilistic neural network based on HMMs known as the recurrent log-linearized Gaussian mixture network (R-LLGMN) 
and showed that the R-LLGMN achieves high classification performance with a smaller amount of training data 
compared with HMMs and conventional NNs. 

Although such model-based NNs have high classification performance, 
they often suffer from problems caused by high dimensionality.
The increased input dimensionality of NNs because of high-dimensional features (e.g., signals measured with numerous electrodes and frequency spectra) 
causes problems such as poor generalization capability, parameter learning difficulty and longer computation time.
These phenomena are called the ``curse of dimensionality'' \cite{bellman1961adaptive}.

To avoid these problems, dimensionality reduction techniques are used before classification 
\cite{guler2005classification,subasi2010eeg,agarwal2010face,mahajan2011classification,hoya2003classification,bu2010eeg,tao2007generalAveraged,tao2007generalTensor,tao2009geometric,zhou2013double,
nie2007neighborhood,nie2009extracting,nie2010flexible,huang2012semi}.
For example, G\"{u}ler {\it et al.} \cite{guler2005classification} applied principal component analysis (PCA) 
to the frequency spectra of electromyograms, 
and classified them using a multi-layer perceptron and a support vector machine. 
Bu {\it et al.} \cite{bu2010eeg} classified time-series data combining linear discriminant analysis (LDA) 
and a recurrent probabilistic neural network.
However, there is no guarantee that the reduced input data can be classified correctly, 
because the dimensionality reduction stage and the classification stage are composed separately.

In contrast, the authors previously proposed a novel time-series pattern classification model called 
time-series discriminant component analysis (TSDCA) \cite{hayashi2013bioelectric}, 
which allows a reduction in the dimensionality of input data using several orthogonal transformation matrices and 
enables the calculation of posterior probabilities for classification
under the assumption that the reduced feature vectors obey an HMM.
We also proposed a probabilistic neural network based on TSDCA 
in which the parameters of TSDCA are obtained as weight coefficients of the NN.
In the learning process of the proposed network, Gram-Schmidt orthonormalization was conducted to maintain the orthogonality 
of the transformation matrices for dimensionality reduction; thus, the convergence of the learning was not guaranteed.

This paper proposes a novel recurrent probabilistic neural network, a time-series discriminant component network (TSDCN), that 
improves the learning algorithm of our previously proposed network. 
In the new learning algorithm, the Lagrange multiplier method is integrated
with a backpropagation through time-based learning process 
to guarantee the convergence of learning while maintaining the orthogonality of the transformation matrices. 
In this way, the TSDCN can obtain the parameters of dimensional reduction and classification simultaneously, 
thereby supporting the accurate classification of time-series data with high dimensionality.

The rest of this paper is organized as follows: Section II describes TSDCA. 
The structure and the learning algorithm of the TSDCN are explained in Section III. 
The verification of the classification ability using high-dimensional artificial data and electroencephalograms (EEGs) 
are presented in Section IV and V. 
Finally, Section VI concludes the paper.

\section{Time-series discriminant component analysis (TSDCA)}
\subsection{Model structure \cite{hayashi2013bioelectric}}
Fig. \ref{TSDCA} shows the overview of TSDCA.
TSDCA consists of several orthogonal transformation matrices and an HMM that incorporates a GMM for the probabilistic density function.
The model allows a reduction in the dimensionality of input data and enables the calculation of posterior probabilities for each class.

In regard to classifying a $D$-dimensional time-series vector $\mbox{\boldmath $x$}(t) \in {\Re}^{D}$ ($t = 1, \cdots, T$) 
into one of the given $C$ classes, the posterior probability $P(c|\mbox{\boldmath $x$}(t))$ ($c=1,\cdots,C$) is examined.
First, $\mbox{\boldmath $x$}(t)$ is projected into a $D'$-dimensional vector ${\mbox{\boldmath $x'$}}^{(c,k,m)}(t) \in {\Re}^{D'}$ using several orthogonal transformation matrices ${{\bf V}^{(c,k,m)}}$. 
This can be described as follows:
\begin{equation}
	{{\mbox{\boldmath $x'$}}^{(c,k,m)}(t)} = {{\bf V}^{(c,k,m)}}^{\rm T}({\mbox{\boldmath $x$}}(t) - {\mbox{\boldmath $\mu$}}^{(c,k,m)}),
	\label{trans}
\end{equation}
where ${\mbox{\boldmath $\mu $}}^{(c,k,m)} \in {\Re}^{D}$ is the mean vector of the component $\{c,k,m\}$ ($k = 1,\cdots,K_c$; $K_c$ is the number of states,
$m = 1,\cdots, M_{c,k}$; $M_{c,k}$ is the number of components), and ${{\bf V}^{(c,k,m)}}\in {\Re}^{D \times D'}$ is the orthogonal transformation matrix that projects from $D$ into $D'$.

In the compressed feature space, the projected data obey a probabilistic density function as follows:
\begin{eqnarray}
	{g({\mbox{\boldmath $x$}}(t);c,\!k,\!m)}
	\!=\! {(2 \pi)}\!^{-\!\frac{D'}{\rm 2}} {|{\Sigma'}^{(c,k,m)}|}^{-\frac{1}{2}}\!\exp[\psi^{(c,k,m)}(t)] \label{pdf},\\
	{\psi^{(c,k,m)}(t)}
	\!=\! -\frac{1}{2}{{\mbox{\boldmath $x'$}}^{(c,k,m)}(t)}\!^{\rm T}{({\Sigma'}^{(c,k,m)})}\!^{-1}\!{\mbox{\boldmath $x'$}}^{(c,k,m)}(t),
	\label{phi}
\end{eqnarray}
where ${\Sigma'}^{(c,k,m)}\in{\Re}^{D' \times D'}$ is the covariance matrix in the compressed feature space. 
\begin{figure}[t]
	\centering
	\includegraphics[width=1.0\hsize] {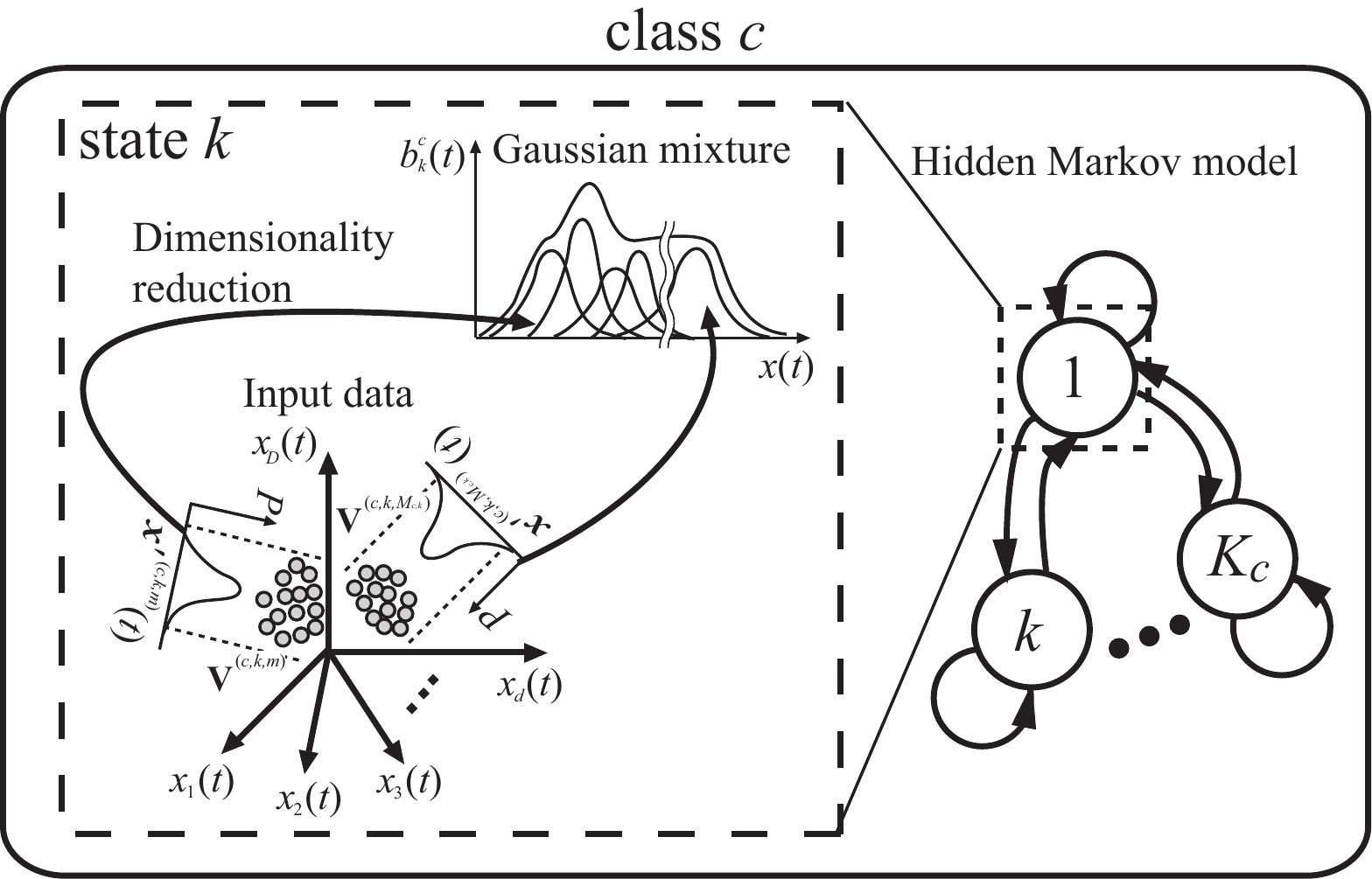}
	\caption{Overview of time-series discriminant component analysis (TSDCA). 
			TSDCA is an expansion of a hidden Markov model (HMM). 
			In this analysis, each class is assumed to be a Markov process with unobserved states. 
			Each state has a Gaussian mixture model (GMM) that incorporates several orthogonal transformation matrices, 
			allowing a reduction in the dimensionality of the time-series data and a calculation of the posterior probabilities 
			of input data for each class.}
	\label{TSDCA}
\end{figure}

Assuming that the projected data obey an HMM, the posterior probability of $c$ given $\mbox{\boldmath $x$}(t)$ is calculated as
\begin{equation}
\hspace{-21mm}
P(c|{\mbox{\boldmath $x$}}(t))
	= \sum_{k=1}^{K_c}\frac{\alpha_k^c(t)}{\sum_{c'=1}^C\sum_{k'=1}^{K_{c'}}\alpha_{k'}^{c'}(t)},
\label{S3C1}
\end{equation}
\begin{equation}
	\hspace{-46mm}
	\alpha_k^c(1) = \pi_k^c b_{k}^c({\mbox{\boldmath $x$}}(1)),
\label{S3C3}
\end{equation}
\vspace{-5mm}
\begin{equation}
	\alpha_k^c(t) = \sum_{k'=1}^{K_c}\alpha_{k'}^c(t-1)\gamma^c_{k',k}b_k^c({\mbox{\boldmath $x$}}(t)) 
	\hspace{2mm} (1<t \leq T),
\label{S3C2}
\end{equation}
where $\gamma^c_{k',k}$ is the probability of a state change from $k'$ to $k$ in class $c$, $b_k^c({\mbox{\boldmath $x$}}(t))$ 
is defined as the likelihood of $\mbox{\boldmath $x$}(t)$ corresponding to the state $k$ in class $c$, and the prior probability $\pi_k^c$ is equal to $P(c,k)|_{t=0}$.
Here, $b_k^c({\mbox{\boldmath $x$}}(t))$ can be derived with the form
\begin{equation}
	b_k^c({\mbox{\boldmath $x$}}(t))
	= \sum_{m=1}^{M_{c,k}}r_{c,k,m}g({\mbox{\boldmath $x$}}(t);c,k,m),
\label{S3C4}
\end{equation}
where $r_{c,k,m}$ represents the mixture proportion. 

Incidentally, TSDCA can be simplified in the particular case where $K_c = 1$ and $T = 1$. 
Since $\gamma^c_{k',k}$ becomes $1$ in this case, $P(c|\bm{x}(1))$ can be calculated instead of (\ref{S3C1}) as 
\begin{equation}
P(c|\bm{x}(1)) = \frac{\pi^c_1\sum_{m=1}^{M_{c,1}}r_{c,1,m}g(\bm{x}(t);c,1,m)}{\sum_{c'=1}^C\pi^c_1\sum_{m=1}^{M_{c,1}}r_{c,1,m}g(\bm{x}(t);c,1,m)}.
\end{equation}
This is equivalent to the posterior probability calculation for static data based on a GMM. 
TSDCA therefore includes classification based on a simple GMM and can be applied to not only time-series data classification but also static data classification.

\subsection{Log-linearization}
An effective way to acquire dimensionality reduction appropriate for classification 
is to train the dimensionality reduction part and classification part together with a single criterion \cite{lotlikar2000bayes}.
In this paper, the authors address this issue by incorporating TSDCA into a neural network 
so that the parameters of TSDCA can be trained simultaneously as network coefficients through a
backpropagation training algorithm.

In preparation for the incorporation, (\ref{trans}) and (\ref{S3C4}) are considered based on a linear combination of coefficient matrices and input vectors.
First, ${\mbox{\boldmath $x'$}}^{(c,k,m)}(t)$ is transformed as follows:
\begin{eqnarray}
\lefteqn{\mbox{{\boldmath $x'$}}^{(c,k,m)}(t)}\nonumber \\ 
&=&{{\bf V}^{(c,k,m)}}^{\rm T}({\mbox{\boldmath $x$}}(t) - {\mbox{\boldmath $\mu$}}^{(c,k,m)})\nonumber\\
&=& {{\bf V}^{(c,k,m)}}^{\rm T}{\mbox{\boldmath $x$}}(t) - \mbox{\boldmath $\mu'$}^{(c,k,m)}\nonumber \\ 
&=& \left[\!\!
\begin{array}{cccc}
-{\mbox{$\mu'$}}_1^{(c,k,m)} & { v_{1,1}^{(c,k,m)}} & \cdots & { v_{D,1}^{(c,k,m)}} \\
\vdots & \vdots & \ddots & \vdots \\
-\mbox{$\mu'$}^{(c,k,m)}_{D'} & { v_{1,D'}^{(c,k,m)}} & \cdots & { v_{D,D'}^{(c,k,m)}} \\
\end{array} 
\!\!\right]\!
\left[\!\!
\begin{array}{c}
1 \\
{{\mbox{\boldmath $x$}}(t)}
\end{array} 
\!\!\right] \nonumber\\
&\triangleq& {{\bf W}^{(c,k,m)}}^{\rm T}
{\bf X}(t),
\label{firstTrans}
\end{eqnarray}
where $\bm{\mu'}^{(c,k,m)}\in{\Re}^{D'}$ corresponds to an image of the mean vector mapped onto the compressed space from the input space.
Hence, ${\mbox{\boldmath $x'$}}^{(c,k,m)}(t)$ is expressed by the multiplication of the coefficient matrix ${\bf W}^{(c,k,m)}$ 
and the novel input vector ${\bf X}(t) = [1, {{\mbox{\boldmath $x$}}(t)}^{\rm T}]^{\rm T}\in {\Re}^{D+1}$.
Secondly, setting 
\begin{equation}
	\xi^{c}_{k',k,m}(t) = \gamma^c_{k',k}r_{c,k,m}g({\mbox{\boldmath $x$}}(t);c,k,m)
\end{equation}
and taking the log-linearization of $\xi^{c}_{k',k,m}(t)$ gives 
\begin{eqnarray}
\lefteqn{\log\xi^{c}_{k',k,m}(t)} \nonumber \\
&=& [\log{\gamma^c_{k',k}\!+\!\log{r_{c,k,m}}}\!-\!\frac{D'}{2}\log{2\pi}\!-\!\frac{1}{2}\log|{\Sigma'}^{(c,k,m)}|,\nonumber\\
& & -\frac{1}{2}{s'}_{1,1}^{(c,k,m)},-{s'}_{1,2}^{(c,k,m)},\cdots ,{s'}_{1,D'}^{(c,k,m)},\cdots,\nonumber\\
& &-\frac{1}{2}(2 - {\delta}_{i,j}){s'}_{i,j}^{(c,k,m)}
\!\cdots\! ,\!-\frac{1}{2}{s'}_{D',D'}^{(c,k,m)}]{\bf X'}^{(c,k,m)}\!(t)\nonumber\\
&\triangleq& {{\bf W'}^{(c,k',k,m)}}^{\rm T}{\bf X'}^{(c,k,m)}(t),
\label{secondTrans}
\end{eqnarray}
where ${s'}_{1,1}^{(c,k,m)},\cdots,{s'}_{D',D'}^{(c,k,m)}$ are elements of the inverse matrix ${({\Sigma'}^{(c,k,m)})}^{-1}$, and ${\delta}_{i,j}$ is a Kronecker delta, which is 1 if $i = j$ and otherwise 0.
Additionally, ${\bf X'}^{(c,k,m)}(t) \in {\Re}^{H}(H = 1+ \frac{D'(D'+1)}{2})$ is defined as
\begin{eqnarray}
\lefteqn{{\bf X'}^{(c,k,m)}(t)}\nonumber\\
&=&[1,{x'^{(c,k,m)}_1(t)}^2,x'^{(c,k,m)}_1(t)x'^{(c,k,m)}_2(t),\cdots ,\nonumber\\
& &x'^{(c,k,m)}_1(t)x'^{(c,k,m)}_{D'}(t), {x'^{(c,k,m)}_2(t)}^2,\nonumber\\
& &x'^{(c,k,m)}_2(t)x'^{(c,k,m)}_3(t), \cdots ,\nonumber \\
& &x'^{(c,k,m)}_2(t)x'^{(c,k,m)}_{D'}(t), \cdots ,{x'^{(c,k,m)}_{D'}(t)}^2]^{\rm T}.
\label{nonlinear}
\end{eqnarray}

As stated above, the parameters of TSDCA can be expressed with a smaller number of coefficients ${\bf W}^{(c,k,m)}$ and ${\bf W'}^{(c,k',k,m)}$ using log-linearization.
If these coefficients are appropriately obtained, the parameters and the structure of the model can be determined and the posterior probability of high-dimensional time-series data for each class can be calculated.

The next section describes how ${\bf W}^{(c,k,m)}$ and ${\bf W'}^{(c,k',k,m)}$ are acquired as weight coefficients of a NN through learning.

\section{Time-series discriminant component network (TSDCN)}
\subsection{Network structure}
Fig. \ref{Structure} shows the structure of the TSDCN,
which is a seven-layer recurrent type with weight coefficients ${\bf W}^{(c,k,m)}$ and ${\bf W'}^{(c,k',k,m)}$ 
between the first/second and third/fourth layers, respectively,
and a feedback connection between the fifth and sixth layers. 
The computational complexity of each layer is proportional to the number of units that is specified below.
\begin{figure*}[t]
	\centering
	\includegraphics[width=1.0\hsize] {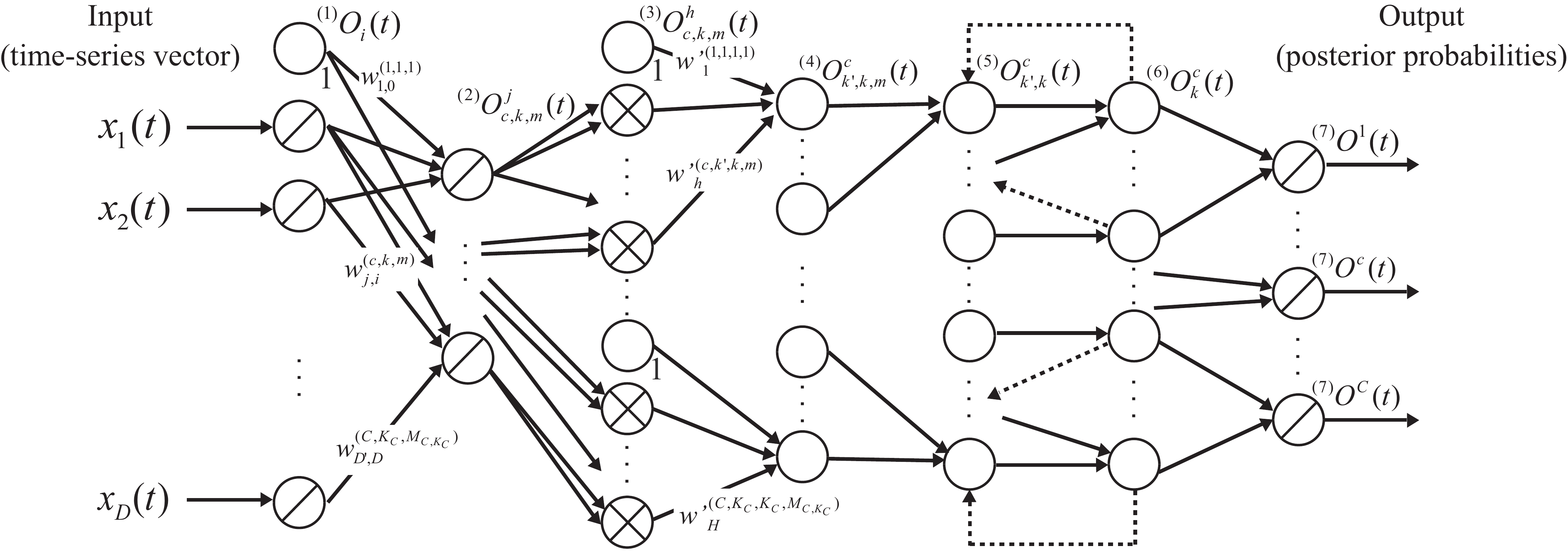}
	\caption{Structure of the time-series discriminant component network (TSDCN). 
				The TSDCN is constructed by incorporating the calculation of TSDCA into the network structure, 
				and consequently consists of seven layers. 
				$\bigcirc$, $\oslash$, and $\otimes$ in the figure stand for a linear sum unit, an identity unit, 
				and a multiplication unit, respectively. 
				The weight coefficients between the first and second layer correspond to 
				the orthogonal transformation matrices and the mean vectors of compressed data, 
				and conduct dimensionality reduction of input data. 
				The weight coefficients between the third and fourth layer includes the probabilistic parameters of GMMs and HMMs.
				The recurrent connections between the fifth and sixth layer correspond to the state changes of HMMs.
				Because of this structure, a calculation equivalent to TSDCA can be implemented.}
	\label{Structure}
\end{figure*}

The first layer consists of $D+1$ units corresponding to the dimensions of the input data ${{\mbox{\boldmath $x$}}(t)}$ $(t = 1,2,\cdots,T) \in {\Re}^{D}$.
The relationships between the input and the output are defined as
\begin{eqnarray}
{}^{(1)}I_i(t) &=& \left\{ 
\begin{array}{l}
1 \:\:\:\:\:\:\:\:(i = 0) \\
x_i(t) \hspace{1mm}(i = 1,\cdots ,D)
\end{array}
\right., \\
{}^{(1)}O_i(t) &=& {}^{(1)}I_i(t),
\end{eqnarray}
where ${}^{(1)}I_i(t)$ and ${}^{(1)}O_i(t)$ are the input and output of the $i$th unit, respectively. 
This layer corresponds to the construction of ${\bf X}(t)$ in (\ref{firstTrans}).

The second layer is composed of $C \times K_c \times M_{c,k} \times D'$ units, 
each receiving the output of the first layer weighted by the coefficient $w_{i,j}^{(c,k,m)}$.
The relationships between the input ${}^{(2)}I^{c,k,m}_j(t)$ and the output ${}^{(2)}O^{c,k,m}_j(t)$ 
of the unit $\{j,c,k,m\}\:(j = 1, \cdots, D'$, $c = 1,\cdots,C$, $k = 1,\cdots,K_c$, $m = 1,\cdots,M_{c,k})$ are described as
\begin{eqnarray}
{}^{(2)}I^{c,k,m}_j(t) &=& \sum^{D}_{i = 0}{}^{(1)}O_i(t)w_{i,j}^{(c,k,m)}, \\
{}^{(2)}O^{c,k,m}_j(t) &=& {}^{(2)}I^{c,k,m}_j(t),
\end{eqnarray}
where the weight coefficient $w_{i,j}^{(c,k,m)}$ is for each element of the matrix ${\bf W}^{(c,k,m)}$ described as follows:
\begin{equation}
{\bf W}^{(c,k,m)} = \left[
\begin{array}{ccc}
w_{0,1}^{(c,k,m)} & \cdots & w_{D,1}^{(c,k,m)} \\
\vdots & \ddots & \vdots \\
w_{0,D'}^{(c,k,m)} & \cdots & w_{D,D'}^{(c,k,m)} \\
\end{array} 
\right]^{\rm T}.
\end{equation}
This layer is equal to the multiplication of ${\bf W}^{(c,k,m)}$ and ${\bf X}(t)$ in (\ref{firstTrans}).

The third layer comprises $C \times K_c \times M_{c,k} \times H$ ($H = 1+ \frac{D'(D'+1)}{2}$) units.
The relationships between the input ${}^{(3)}I^{c,k,m}_h(t)$ and the output ${}^{(3)}O^{c,k,m}_h(t)$ of the units $\{h,c,k,m\}\:(h=1,\cdots,H)$ are defined as
\begin{equation}
{}^{(3)}I^{c,k,m}_h(t)
\!=\!\left\{ 
\begin{array}{l}
\hspace{-1mm}1 
\hspace{2mm}(h = 1) \\ 
\hspace{-1mm}{}^{(2)}O^{c,k,m}_j(t){}^{(2)}O^{c,k,m}_{j'}(t)\\ 
\hspace{2mm}(h \!=\! j'\!-\!\frac{1}{2}j^2\!+\!(D'\!+\!\frac{1}{2})j\!-\!D'\!+\!1)
\end{array}
\right.\!\!,
\label{2in2out}
\end{equation}
\begin{equation}
\hspace{-47mm}
{}^{(3)}O^{c,k,m}_h(t) \!=\! {}^{(3)}I^{c,k,m}_h(t),
\end{equation}
where $j \leq j'$($j'$ = 1,$\cdots$,$D'$), and (\ref{2in2out}) corresponds to the nonlinear conversion shown in (\ref{nonlinear}).

The fourth layer comprises $C \times K_c^2 \times M_{c,k}$ units.
Unit $\{c,k',k,m\}(k' = 1,\cdots,K_c)$ receives the output of the third layer weighted by the coefficient ${w'}_h^{(c,k',k,m)}$.
The input ${}^{(4)}I^c_{k',k,m}(t)$ and the output ${}^{(4)}O^c_{k',k,m}(t)$ are defined as
\begin{eqnarray}
{}^{(4)}I^c_{k',k,m}(t) &=& \sum_{h=1}^H {}^{(3)}O^{c,k,m}_h(t) {w'}_h^{(c,k',k,m)},\label{in4}\\
{}^{(4)}O^c_{k',k,m}(t) &=& \exp \left({}^{(4)}I^c_{k',k,m}(t)\right),\ \ \ \ \ \ \ \ \ \ 
\label{out4}
\end{eqnarray}
where the weight coefficient ${w'}_h^{(c,k',k,m)}$ corresponds to each element of the vector ${\bf W'}^{(c,k',k,m)}$.
\begin{equation}
{\bf W'}^{(c,k',k,m)} = \left[
\begin{array}{ccc}
\!\!\!{w'}_1^{(c,k',k,m)}, & \!\!\!\!\!\cdots, & \!\!\!\!\!{w'}_H^{(c,k',k,m)\!\!\!}
\end{array} 
\right]^{\rm T}.
\end{equation}
(\ref{in4}) stands for the multiplication of ${\bf W'}^{(c,k',k,m)}$ and ${\bf X'}^{(c,k,m)}(t)$ in (\ref{secondTrans}). 
(\ref{out4}) corresponds to the exponential function of a Gaussian distribution in (\ref{pdf}).

The fifth layer consists of $C \times K_c^2$ units.
The output of the fourth layer is added up and input into this layer.
The one-time-prior output of the sixth layer is also fed back to the fifth layer. 
These are expressed as follows:
\begin{eqnarray}
{}^{(5)}I^c_{k',k}(t) &=& \sum_{m=1}^{M_{c,k}}{}^{(4)}O^c_{k',k,m}(t),\label{in5} \\
{}^{(5)}O^c_{k',k}(t) &=& {}^{(6)}O^c_{k'}(t-1){}^{(5)}I^c_{k',k}(t),\ \ 
\label{out5}
\end{eqnarray}
where ${}^{(6)}O^c_{k'}(0)=1.0$ for the initial phase. 
This layer represents the summation over components in (\ref{S3C4})
and the multiplication of $\alpha_{k'}^c(t-1)$, $\gamma^c_{k',k}$, and $b_k^c({\mbox{\boldmath $x$}}(t))$ in (\ref{S3C2}).

The sixth layer has $C \times K_c$ units.
The relationships between the input ${}^{(6)}I^c_k(t)$ and the output ${}^{(6)}O^c_k(t)$ of the unit \{$c,k$\} are described as
\begin{eqnarray}
{}^{(6)}I^c_k(t) &=& \sum_{k'=1}^{K_c}{}^{(5)}O^c_{k',k}(t),\\
{}^{(6)}O^c_k(t) &=& \frac{{}^{(6)}I^c_k(t)}{\sum_{c'=1}^C\sum_{k'=1}^{K_{c'}}{}^{(6)}I^{c'}_{k'}(t)}.
\end{eqnarray}
This layer corresponds to the summation over states in (\ref{S3C2}) and the normalization in (\ref{S3C1}).

Finally, the seventh layer consists of $C$ units, and its input ${}^{(7)}I^c(t)$ and output ${}^{(7)}O^c(t)$ are 
\begin{eqnarray}
{}^{(7)}I^c(t) &=& {\sum_{k=1}^{K_c}}{}^{(6)}O^c_k(t),\\
{}^{(7)}O^c(t) &=& {}^{(7)}I^c(t).\ \ \ \ \ \ \ \ \ \ \ \ \ \ \ \ \ \ \ 
\end{eqnarray}
${}^{(7)}O^c(t)$ corresponds to the posterior probability for class $c$ $P(c|{\mbox{\boldmath $x$}}(t))$.
Here, the posterior probability $P(c|{\mbox{\boldmath $x$}}(t))$ based on TSDCA can be calculated if the NN coefficients ${\bf W}^{(c,k,m)}$, ${\bf W'}^{(c,k',k,m)}$ are appropriately established.

\subsection{Learning algorithm}
The learning process of the TSDCN is required to achieve maximization of the likelihood and 
orthogonalization of the transformation matrices ${\bf V}^{(c,k,m)}$, simultaneously. 
In our previous paper, the weight modulations based on the maximum likelihood and the Gram-Schmidt process for orthogonalization 
were conducted alternately \cite{hayashi2013bioelectric}. 
Unfortunately, the Gram-Schmidt process interfered with the monotonic increase of likelihood, 
hence the convergence of network learning was not theoretically guaranteed. 

This paper addresses this issue by introducing the Lagrange multiplier method to the learning algorithm. 
In preparation, let us redefine ${\bf W}^{(c,k,m)}$ as follows:
\begin{equation}
{\bf W}^{(c,k,m)}
\!\!=\!\! \left[\!\!\!
\begin{array}{cccc}
{-{\bm{\mu'}}^{(c,k,m)}}^{\rm T}\!, &\!\! {\bm{v}_1^{(c,k,m)}}^{\rm T}\!,\! &\!\! \cdots\!, \!&\!\! {\bm{v}_{D'}^{(c,k,m)}}^{\rm T} \\
\end{array} 
\!\!\!\right]^{\rm T}\!\!\!,\!
\end{equation}
where $\bm{v}_j^{(c,k,m)}~(j = 1, \cdots, D')$ is the $j$th column vector of ${\bf V}^{(c,k,m)}$. 
A set of input vectors ${\mbox{\boldmath $x$}}^{(n)}(t)$ ($n = 1, \cdots, N$; $N$ is the number of training data) 
is given for training with the teacher vector 
${\bm Q}^{(n)} = {[Q^{(n)}_1,\cdots,Q^{(n)}_c,\cdots,Q^{(n)}_C]}^{\rm T}$ for the $n$th input, 
where $Q^{(n)}_c = 1$ and $Q^{(n)}_{c'} = 0$ ($c' \neq c$) for the training sample of class $c$. 
Now consider the following optimization problem: 
\begin{eqnarray}
{\rm minimize} \quad \hspace{2mm} J\hspace{48mm}\label{optProb1}\nonumber \\
{\rm subject~to} \quad h^{(c,k,m)}_i = 0 \quad (i = 1, \cdots, N_{\rm const}).
\end{eqnarray}
Here, $J$ is a negative log-likelihood function described as 
\begin{equation}
J = \sum^N_{n = 1}J_n = -\sum^{N}_{n = 1}\sum^{C}_{c = 1}Q^{(n)}_c\log{}^{(7)}O^{c}(T)^{(n)}, 
\label{J}
\end{equation} 
where ${}^{(7)}O^{c}(T)^{(n)}$ is the output for the $n$th input at $T$.
$h^{(c,k,m)}_i$ is a constraint that represents orthogonality conditions defined as
\begin{eqnarray}
	h^{(c,k,m)}_i &=& {\bm{v}^{(c,k,m)}_j}^{\rm T}{\bm{v}^{(c,k,m)}_l} - \delta_{j,l} \nonumber \\
				  & & (i = \frac{1}{2}(l-1)l + j, 1\leq j \leq l \leq D'),
\end{eqnarray}
$N_{\rm const}$ is the number of constraints given as $D'(D'+1)/2$. 
Using the method of Lagrange multiplier, the constrained optimization problem can be converted to an unconstrained problem defined as 
\begin{equation}
L = J + \sum_{i=1}^{N_{\rm const}}{\lambda}^{(c,k,m)}_i h^{(c,k,m)}_i,\label{optProb2}
\end{equation}
where ${\lambda}^{(c,k,m)}_i$ is a Lagrange multiplier. 
By differentiating $L$ with respect to ${\bf W}^{(c,k,m)}$, ${\bf W'}^{(c,k',k,m)}$, and ${{\bm \lambda}}^{(c,k,m)}$, 
the necessary conditions of the above optimization problem are derived as
\begin{eqnarray}
	\nabla \bm{J} + \nabla {\bm{h}^{(c,k,m)}}^{\rm T} {{\bm \lambda}}^{(c,k,m)} &=& {\bf 0}, \label{necOpt1}\\
	\bm{h}^{(c,k,m)} &=& {\bf 0} \label{necOpt2},
\end{eqnarray}
where $\nabla$ is a differential operator with respect to ${\bf W}^{(c,k,m)}$ and ${\bf W'}^{(c,k',k,m)}$.
The log-likelihood function $J$ can therefore be minimized while maintaining orthogonality 
by obtaining a point that satisfies (\ref{necOpt1}) and (\ref{necOpt2}) 
using an arbitrary nonlinear optimization algorithm.

Now we consider the minimization of $L$ using the gradient method.
The partial differentiations of $L$ with respect to each weight coefficient are given as
\begin{equation}
\frac{\partial L}{\partial {\bf W}^{(c,k,m)}}
		= \frac{\partial J}{\partial {\bf W}^{(c,k,m)}} 
			\!+\! \sum_{i=1}^{N_{\rm const}}{\lambda}^{(c,k,m)}_i \frac{\partial h^{(c,k,m)}_i }{\partial {\bf W}^{(c,k,m)}},
		\label{parL1}
\end{equation}
\begin{equation}
\hspace{-37mm}
\frac{\partial L}{\partial {\bf W'}^{(c,k',k,m)}}
		= \frac{\partial J}{\partial {\bf W'}^{(c,k',k,m)}},
		\label{parL2} 
\end{equation}
thus $L$ can be minimized without constraint with respect to ${\bf W'}^{(c,k',k,m)}$. 
The weight modification for ${\bf W'}^{(c,k',k,m)}$ is then defined as 
\begin{equation}
\label{w2}
\Delta{\bf W'}^{(c,k',k,m)} = -\gamma\sum^N_{n = 1}\frac{\partial J_n}{\partial {\bf W'}^{(c,k',k,m)}},
\end{equation}
where $\gamma$ is the learning rate. 
If the learning rate is appropriately chosen, the convergence of the gradient method is guaranteed.

With regard to ${\bf W}^{(c,k,m)}$, 
let ${{}^{(1)}{\bf W}}^{(c,k,m)}_l$ be an arbitrary ${\bf W}^{(c,k,m)}$ that satisfies (\ref{necOpt2}), 
and the differential vector ${\bm{d}}_l$ to the next point ${{}^{(1)}{\bf W}}^{(c,k,m)}_{l+1}$ is given as 
\begin{eqnarray}
	{\bm{d}}^{(c,k,m)}_l &=& -\nabla \bm{L}({{}^{(1)}{\bf W}}^{(c,k,m)}_l, {{\bm \lambda}}^{(c,k,m)}_l) \nonumber \\
						 &=& -\nabla \bm{J} - \nabla {\bm{h}^{(c,k,m)}_l}^{\rm T} {{\bm \lambda}}^{(c,k,m)}_l, 
	\label{diffvec}
\end{eqnarray}
where $\nabla \bm{J}$ is calculated as 
\begin{equation}
\label{w1}
\nabla \bm{J} = \sum^N_{n = 1}\frac{\partial J_n}{\partial {\bf W}^{(c,k,m)}}.
\end{equation}

Here, the Taylor expansion for (\ref{necOpt2}) at ${{}^{(1)}{\bf W}}^{(c,k,m)}_l$ is 
\begin{eqnarray}
\bm{h}^{(c,k,m)} 
 &=& \bm{h}^{(c,k,m)}_l + (\nabla {\bm{h}^{(c,k,m)}_l}^{\rm T})^{\!\rm T} \nonumber \\
 & & \cdot ({\bf W}^{(c,k,m)} - {{}^{(1)}{\bf W}}^{(c,k,m)}_l).
	\label{tayEx}
\end{eqnarray}
Let us denote a solution of the above equation by ${{}^{(1)}{\bf W}}^{(c,k,m)}_{l+1}$, thereby (\ref{tayEx}) can be rewritten as
\begin{equation}
	\bm{h}^{(c,k,m)}_l + (\nabla {\bm{h}^{(c,k,m)}_l}^{\rm T})^{\rm T} {\bm{d}}^{(c,k,m)}_l = {\bf 0}.
	\label{tayEx2}
\end{equation}
Hence (\ref{diffvec}) and (\ref{tayEx2}) can be combined as follows:
\begin{equation}
\left[\!\!\!
\begin{array}{cc}
I \!\!\!&\!\!\! \nabla {\bm{h}^{(c,k,m)}_l}^{\rm T} \\
(\nabla {\bm{h}^{(c,k,m)}_l}^{\rm T})^{\rm T} \!\!\!&\!\!\! {\bf 0}
\end{array} 
\!\!\!\right]\!\!
\!\left[\!\!\!
\begin{array}{c}
{\bm{d}}^{(c,k,m)}_l \\
{{\bm \lambda}}^{(c,k,m)}_l
\end{array} 
\!\!\!\right]
\!\!=\!\!
\left[\!\!\!
\begin{array}{c}
-\nabla \bm{J} \\
-\bm{h}^{(c,k,m)}_l
\end{array} 
\!\!\!\right]\!,
\end{equation}
where $I$ is an identity matrix.
Solving the above simultaneous equations, ${\bf W}^{(c,k,m)}$ can be modulated using ${\bm{d}}^{(c,k,m)}_l$ as follow: 
\begin{equation}
\Delta{\bf W}^{(c,k,m)} = \gamma {\bm{d}}^{(c,k,m)}_l.
\end{equation}
The detailed calculation of gradient vectors will be described in Appendix.

Using this algorithm, collective training can be applied in relation to the weight coefficients for dimensional reduction and discrimination, 
simultaneously.

\section{Simulation experiments}
To investigate the characteristics of the TSDCN, artificial data classification experiments were conducted. 
The purposes of these experiments were as follows: 
A: To reveal the classification ability of the TSDCN under ideal conditions 
and the influence of network parameter variation, 
B: To show that the TSDCN can reduce training time while maintaining classification accuracy equivalent to that of conventional NNs, 
C: To compare the dimensionality reduction ability of the TSDCN with conventional methods, and 
D: To compare the TSDCN with our previous network.

In terms of the examination of network parameters, Tsuji {\it et al.} had already discussed 
the influence of parameters derived from an HMM such as the number of states $K_c$, 
the number of components $M_{c,k}$, the number of training data $N$, and time-series length $T$ in the R-LLGMN \cite{tsuji2003recurrent}.
Subsection A will then examine the influence of the number of input dimensions $D$ and 
the number of reduced dimensions $D'$ since they make the greatest difference between the TSDCN and the R-LLGMN. 
The variation of the number of classes $C$ will also be examined because it is the most important issue for classification.

In subsection B, the training time of the TSDCN will be compared with those of the R-LLGMN \cite{tsuji2003recurrent} 
and the Elman network \cite{elman1990finding} that are without dimensionality reduction.

The authors will show the differences in the dimensionality reduction characteristics 
between the TSDCN and PCA, and between the TSDCN and LDA using two kinds of visible problems, 
and then verify how the differences influence high-dimensional data classification in subsection C. 

Finally, comparison with our previously proposed network will be conducted in subsection D. 
Since the TSDCN is an improvement over our previous network \cite{hayashi2013bioelectric} with respect to the learning algorithm, 
this comparison demonstrates that the improvement really works better in terms of accuracy and convergency.

The computer used in each experiment was an Intel Core(TM) i7-3770K (3.5 GHz), 16.0 GB RAM.
Furthermore, a terminal attractor was introduced to curtail the training time \cite{zak1989terminal}.

\subsection{Classification under ideal conditions and network parameter variation}
\subsubsection{Method}
\begin{figure}[t]
	\centering
	\includegraphics[width=1.0\hsize] {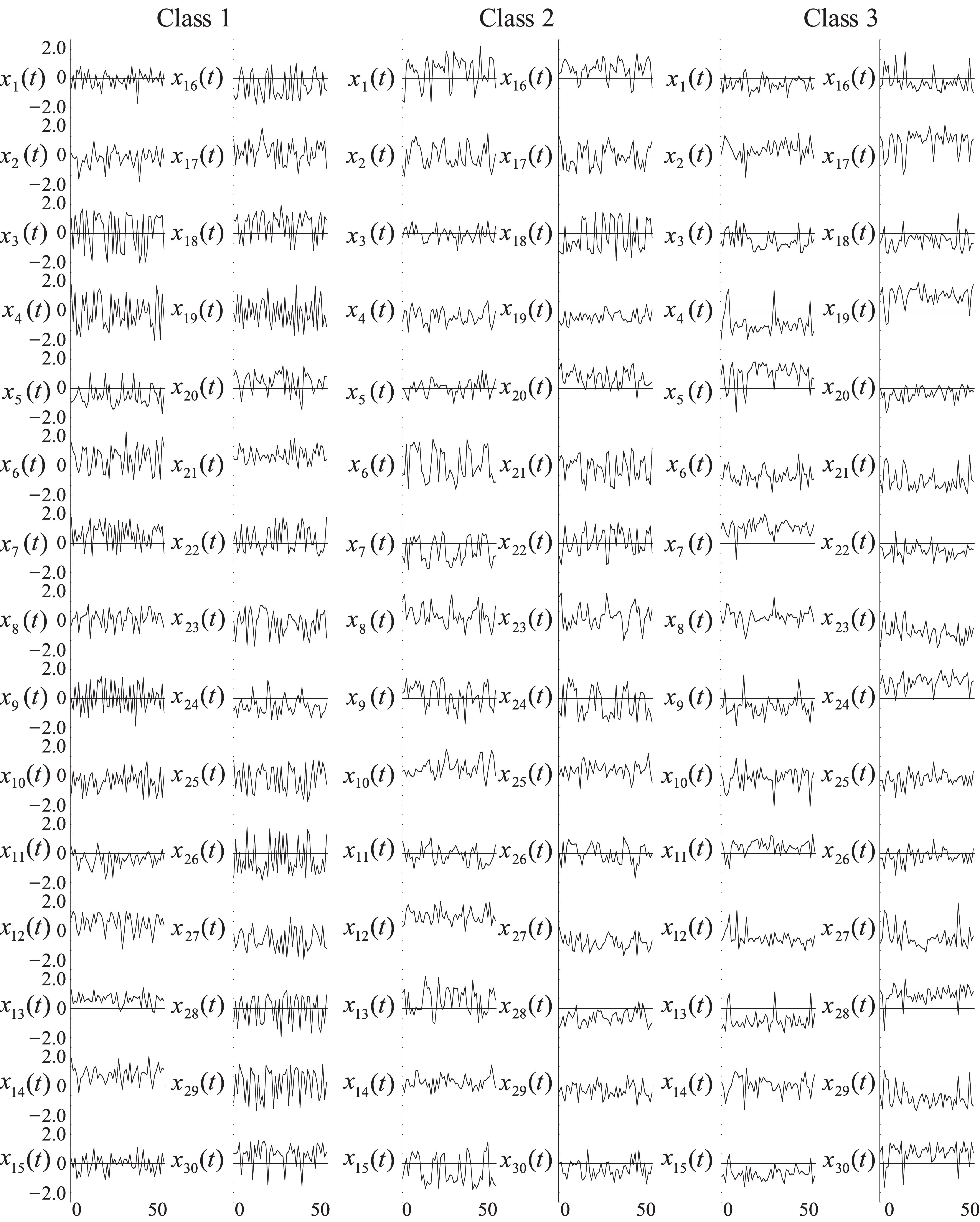}
	\caption{Example of ${\bm x}(t)$ used in the classification experiment. 
			The data were generated based on an HMM for each class  ($C=3$, $D=30$).}
	\label{30dim}
\end{figure}
The data used in the experiment ${\bm x} (t) \in {\Re}^{D}$ were generated based on an HMM for each class 
because the ideal conditions for the TSDCN are that the input signals obey HMMs. 
The HMMs used were fully connected, which had two states and two components for each state. 
The length of the time series $T$ was set as $T=50$.
Fig. \ref{30dim} shows an example of ${\bm x} (t)$.
The parameters used in this figure were $C=3$, $D=30$.
The vertical axis and the horizontal axis in the figure indicate the value of each dimension and time, respectively. 
In terms of the HMM parameters used in the generation of ${\bm x} (t) $, 
the mean vectors and the covariance matrices were set using uniform random numbers in $[-1, 1]$; 
the mixture proportion, the initial state probabilities, 
and the state change probabilities are determined by uniform random vectors normalized in $[0, 1]$. 
In the validation of classification accuracy, 5 samples were treated as training data, 
and 50 samples were used as test data for each class. 
The classification accuracy is defined as 
$100 \times N_{\rm correct}/N_{\rm total}$,
where $N_{\rm correct}$ is the amount of test data correctly classified, 
and $N_{\rm total}$ is the total amount of test data.
Average classification accuracy and training time were then calculated by changing the HMM parameters, 
regenerating the training/test data sets 10 times, 
and resetting the initial weight coefficients of the TSDCN 10 times for each data set.
For the fixed number of states $K_c=2$ and the number of components $M_{c,k}=2$, 
the number of input dimensions $D$ and the number of reduced dimensions $D'$ were varied as 
$D=10, 30, 50, 70, 90$ and $D'=1, 2, 3, 4, 5$, respectively.
The number of classes $C$ was also varied as $C=2, 3, 4, 5, 10, 20$ for fixed $D=30$ and $D'=1$.

\subsubsection{Results}
\begin{figure}[t]
	\centering
	\includegraphics[width=1.0\hsize] {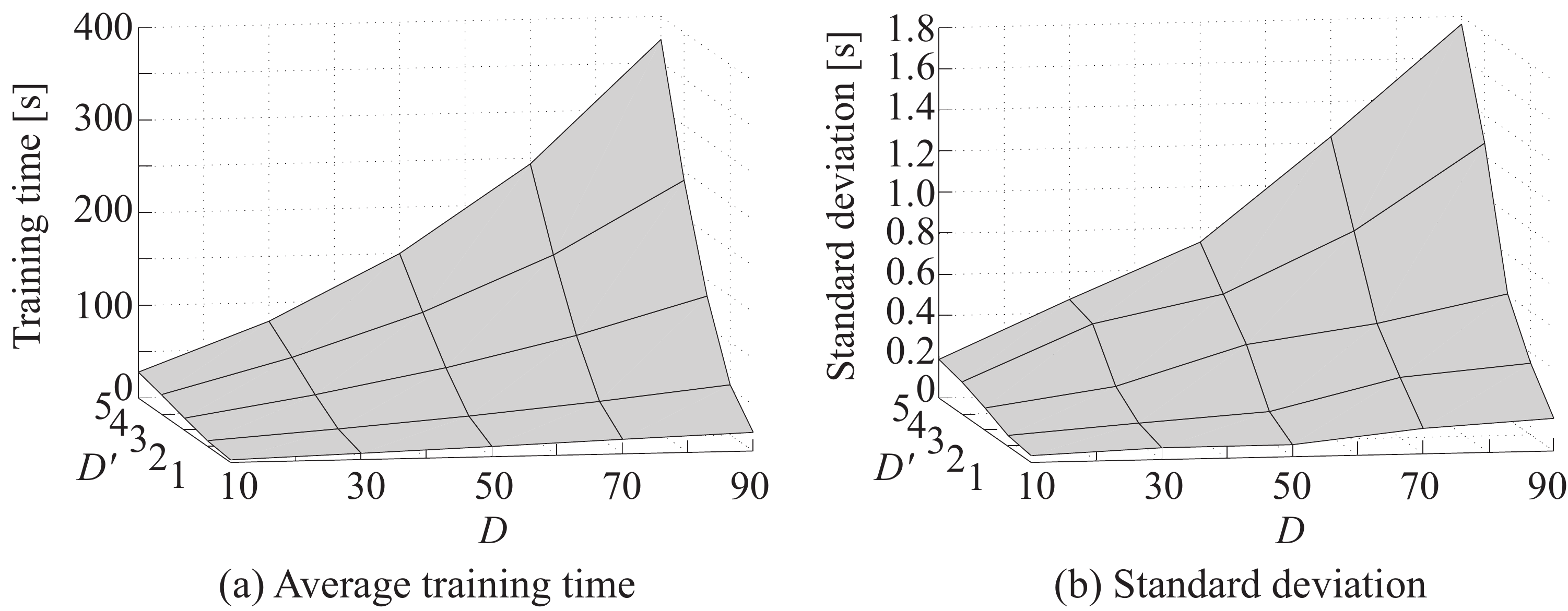}
	\caption{Average training time and standard deviation for each combination of the number of input dimensions $D$ 
			and the number of reduced dimensions $D'$ ($C=3$).}
	\label{dimChange}
\end{figure}
Fig. \ref{dimChange} shows average training time and its standard deviation for each combination of 
the number of input dimensions $D$ and the number of reduced dimensions $D'$ ($C=3$). 
In this experiment, the classification accuracies were $100$ [\%] for any combinations of $D$ and $D'$.

\begin{figure}[t]
	\centering
	\includegraphics[width=1.0\hsize] {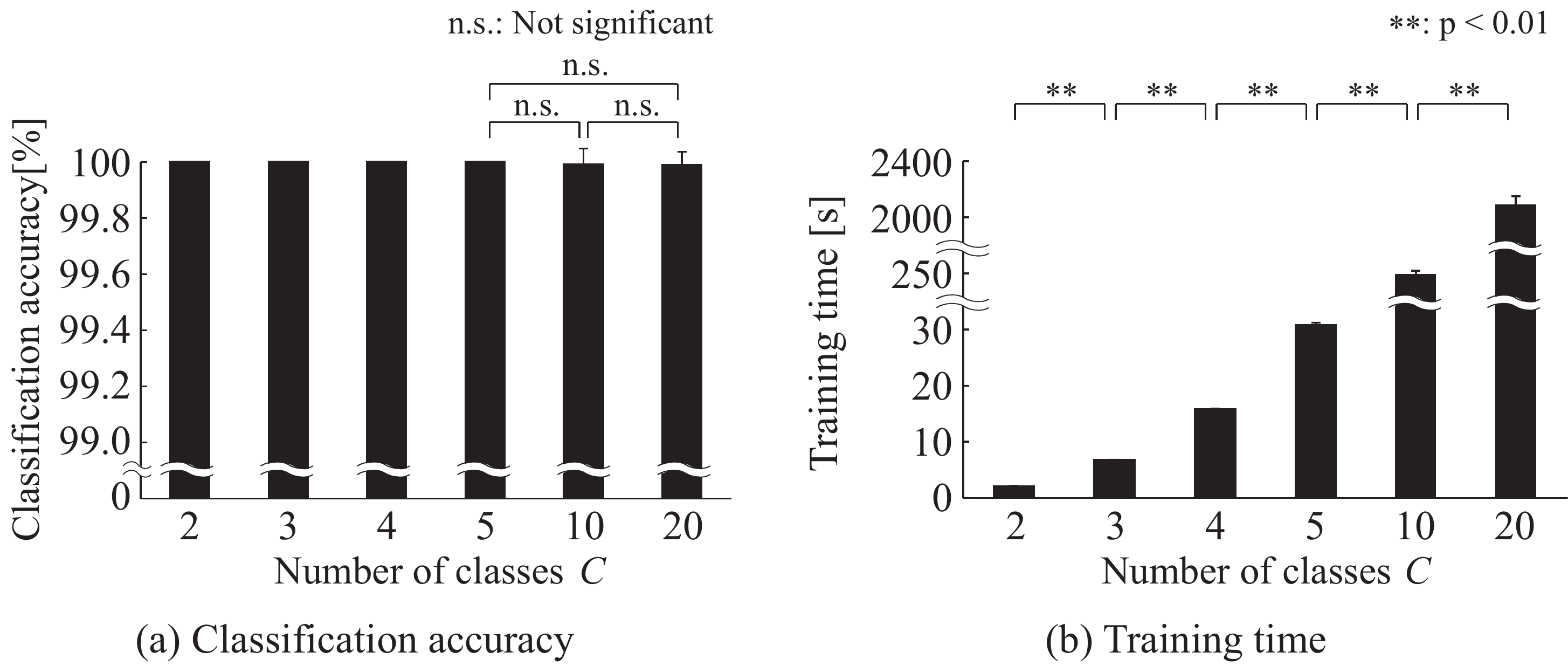}
	\caption{Average classification accuracy and training time of the TSDCN for each number of classes $C$ ($D=30$, $D'=1$). 
			Note that the vertical axis scale from 0 to 99.0 was omitted in (a).}
	\label{classChange}
\end{figure}
Fig. \ref{classChange} shows average classification accuracy and training time for each number of classes $C$.
The classification accuracies for $C=2$ to $C=5$ were $100$ [\%]. 
Although the classification accuracies for $C=10$ and $C=20$ were uneven 
($100.0 \pm 5.6 \times 10^{-2}$ [\%], $100.0 \pm 4.7\times 10^{-2}$ [\%], respectively), 
there were no significant differences among $C=5$, $C=10$, and $C=20$ based on the Holm method \cite{holm1979simple}.

\subsubsection{Discussion}
\begin{table}[t]
	\caption{Calculation complexity of each calculation part}
	\label{order}
	\centering
	\begin{tabular}{c||c|c|c} \hline
		&	Forward calculation	&	Backpropagation	&	Lagrange multiplier\\	\hline \hline
	$D$	&		$\mathcal{O}(D)$			&		$\mathcal{O}(D)$			&		$\mathcal{O}(D^2)$	\\	\hline
	$D'$&		$\mathcal{O}(D'^2)$		&		$\mathcal{O}(D'^3)$		&		$\mathcal{O}(D'^4)$	\\	\hline
	\end{tabular}
\end{table}
In Fig. \ref{dimChange} (a), the increase of the training time according to the increase of $D$ is almost linear 
with a small number of $D'$ such as $D'=1$ and $D'=2$. 
In contrast, the training time increases as a quadratic function with a larger number of $D'$. 
This can be explained on the basis of calculation complexity.
Table \ref{order} summarizes approximate complexity in each part of the calculation. 
When the number of reduced dimensions $D'$ is smaller, the calculation amounts of the forward calculation and the backpropagation 
are substantially larger than in the Lagrange multiplier method, 
hence the total calculation complexity according to the increase of $D$ can be regarded as linear. 
If $D'$ is larger, however, the calculation complexity of the Lagrange multiplier method rapidly increases, 
and the quadratic term of $D$ cannot be ignored.
In other words, the TSDCN can suppress the increase of the amount of calculation according to the increase of the input dimensions 
to linear if the number of reduced dimensions $D'$ is sufficiently small, 
whereas conventional methods such as HMMs and the R-LLGMN have the complexity of $\mathcal{O}(D^2)$ because 
they are required to calculate the covariance matrices of the input vectors.

In the relationship between the classification accuracy and the number of classes $C$ shown in Fig. \ref{classChange}, 
the TSDCN achieved exceedingly high classification accuracy with any $C$ 
although the training time increased as $C$ increased. 
These results indicate that the TSDCN can classify high-dimensional time-series data significantly accurately 
if the assumption that the input data obey HMMs is satisfied. 
In addition, the TSDCN achieved high classification performance even for high multi-class problems such as $C=20$ 
although the number of reduced dimensions was very small ($D'=1$). 
This is because that the TSDCN has orthogonal transformation matrices different for each class, 
hence mappings separate from each class onto reduced-dimensional spaces that are appropriate for classification can be acquired. 
Therefore, there is no need to use a large number of $D'$, 
thus the previously mentioned restriction that $D'$ should be sufficiently small is easily satisfied.

\begin{figure}[t]
	\centering
	\includegraphics[width=1.0\hsize] {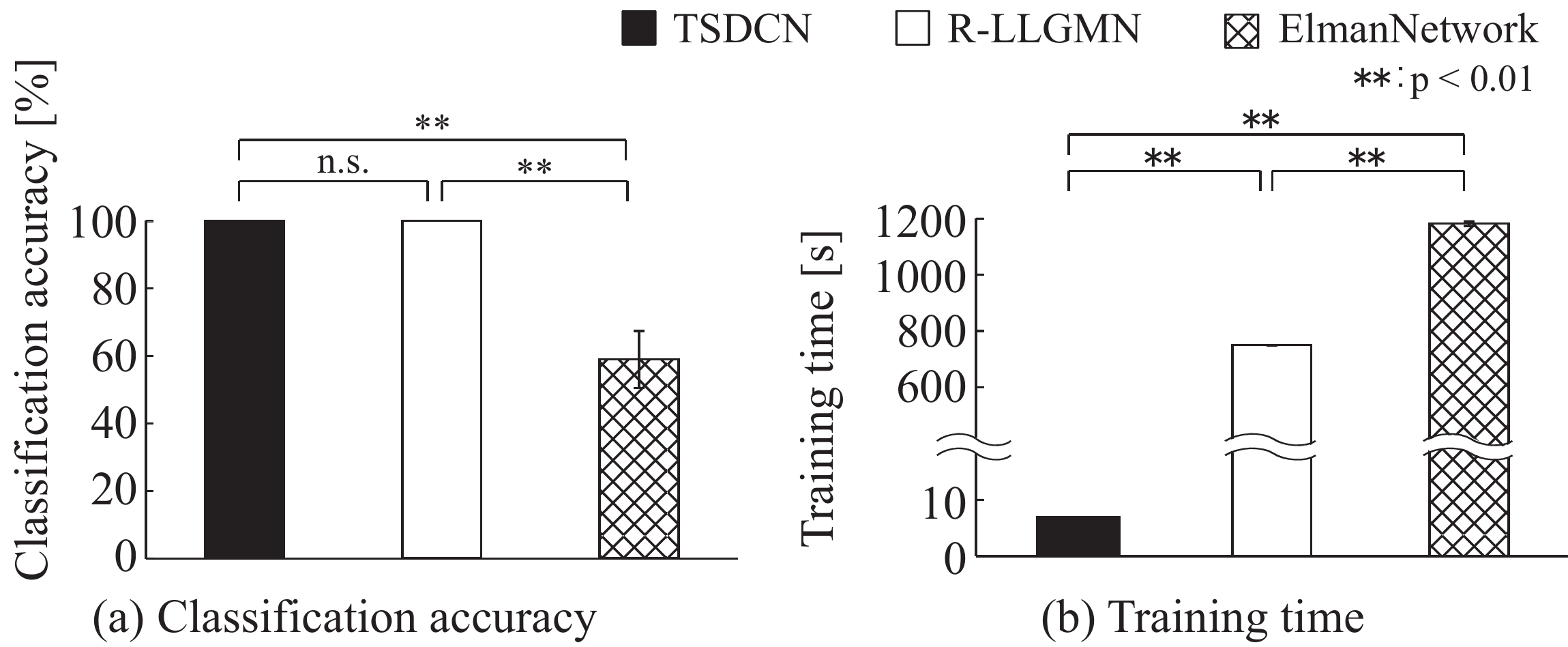}
	\caption{Average classification accuracy and training time of each method for comparison with conventional NNs.}
	\label{AccuracyNonComp}
\end{figure}
\subsection{Comparison with conventional NNs}
\subsubsection{Method}
The data used in this experiment were also generated by an HMM ($C=3$, $D=30$, $T=50$). 
The classification accuracy and training time for each NN were calculated in the same manner as subsection A.
The parameters of the TSDCN were $K_c=2$, $M_{c,k}=2$, $D'=1$, 
and the number of states and components of the R-LLGMN were identical to those of the TSDCN. 
The Elman network had three layers (one hidden) and 30, 10, 3 units for input, hidden, and output layers, respectively. 
The input/output function was a log-sigmoid function, 
and the backpropagation with 0.01 of learning rates and 0.01 of error thresholds was used for training. 

\subsubsection{Results}
Fig. \ref{AccuracyNonComp} shows the classification accuracy and the training time for each method. 
The classification accuracies of both the TSDCN and the R-LLGMN were $100$ [\%]. 
The accuracy of the Elman network was $58.9 \pm 8.4$ [\%], 
and significant differences between the other methods were confirmed. 
The training times of the TSDCN, the R-LLGMN, and the Elman network were 
$6.8 \pm 0.1$ [s], $749.3 \pm 18.2$ [s], and $1181.7 \pm 6.8$ [s], respectively, 
and there were significant differences among all the methods.

\subsubsection{Discussion}
It can be considered that the Elman network estimated the distribution of the high-dimensional input data in the original feature space, 
and made its structure complex, resulting in the low classification accuracy and the significantly long training time.
The R-LLGMN achieved high classification performance since it is developed based on HMMs.
However, the training time was long because the R-LLGMN modeled all the input dimensions using numerous parameters. 
The training time of the TSDCN was significantly reduced 
in spite of the high classification accuracy. 
These results showed that the TSDCN can reduce training time, maintaining the high classification performance.

\subsection{Comparison with conventional dimensionality reduction}
\subsubsection{Method}
\begin{figure}[t]
	\centering
	\includegraphics[width=1.0\hsize] {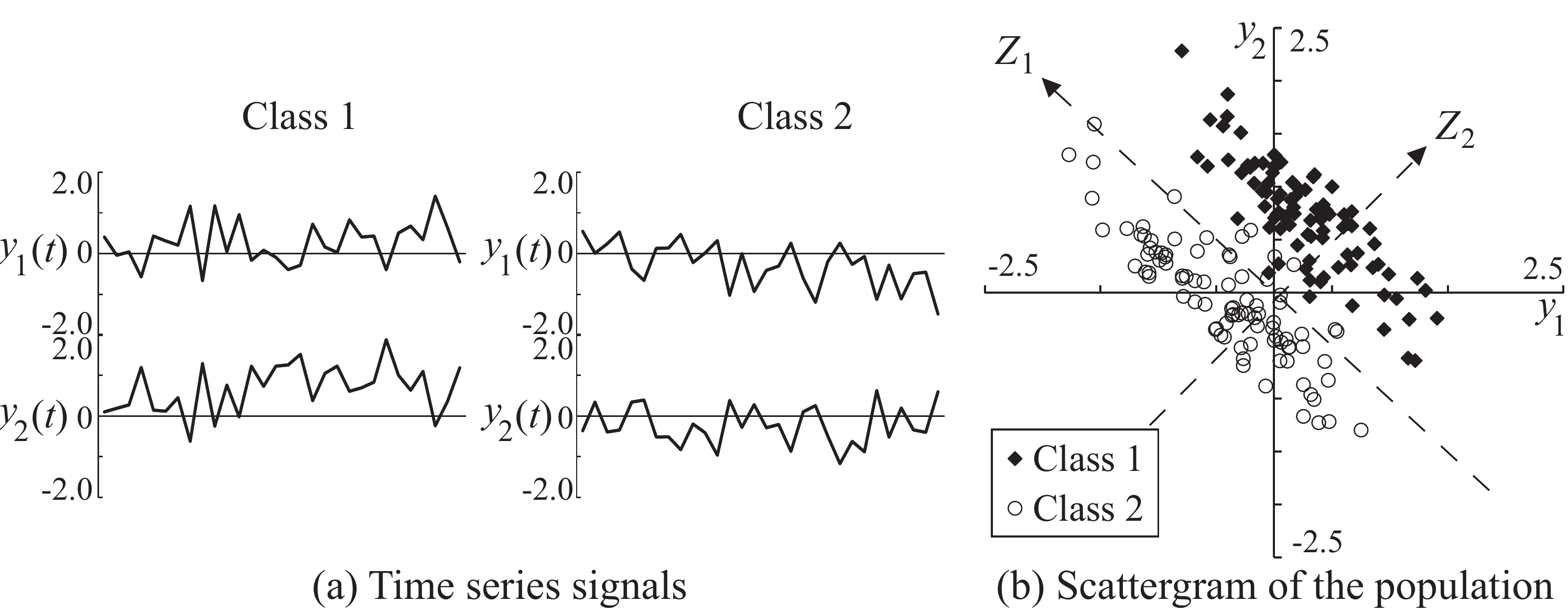}
	\caption{Example of the problem in comparison with PCA. (a) shows an example of time-series signal of each dimension, 
			and (b) indicates a scattergram of the distribution of the population used in the data generation. 
			The auxiliary axis $Z_1$ and $Z_2$ show the first and second principal component, respectively.}
	\label{PCAprob}
\end{figure}
\begin{figure}[t]
	\centering
	\includegraphics[width=1.0\hsize] {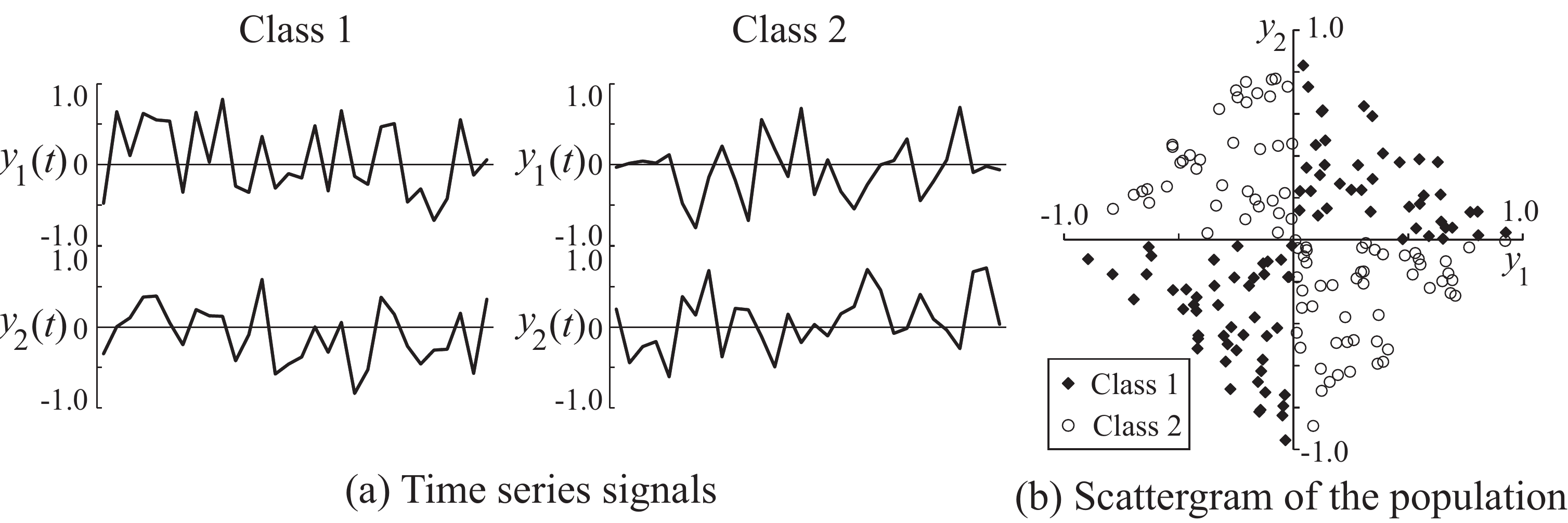}
	\caption{Example of the problem in comparison with LDA. The scattergram (b) shows that the problem is nonlinearly separative.}
	\label{LDAprob}
\end{figure}
\begin{figure}[t]
	\centering
	\includegraphics[width=1.0\hsize] {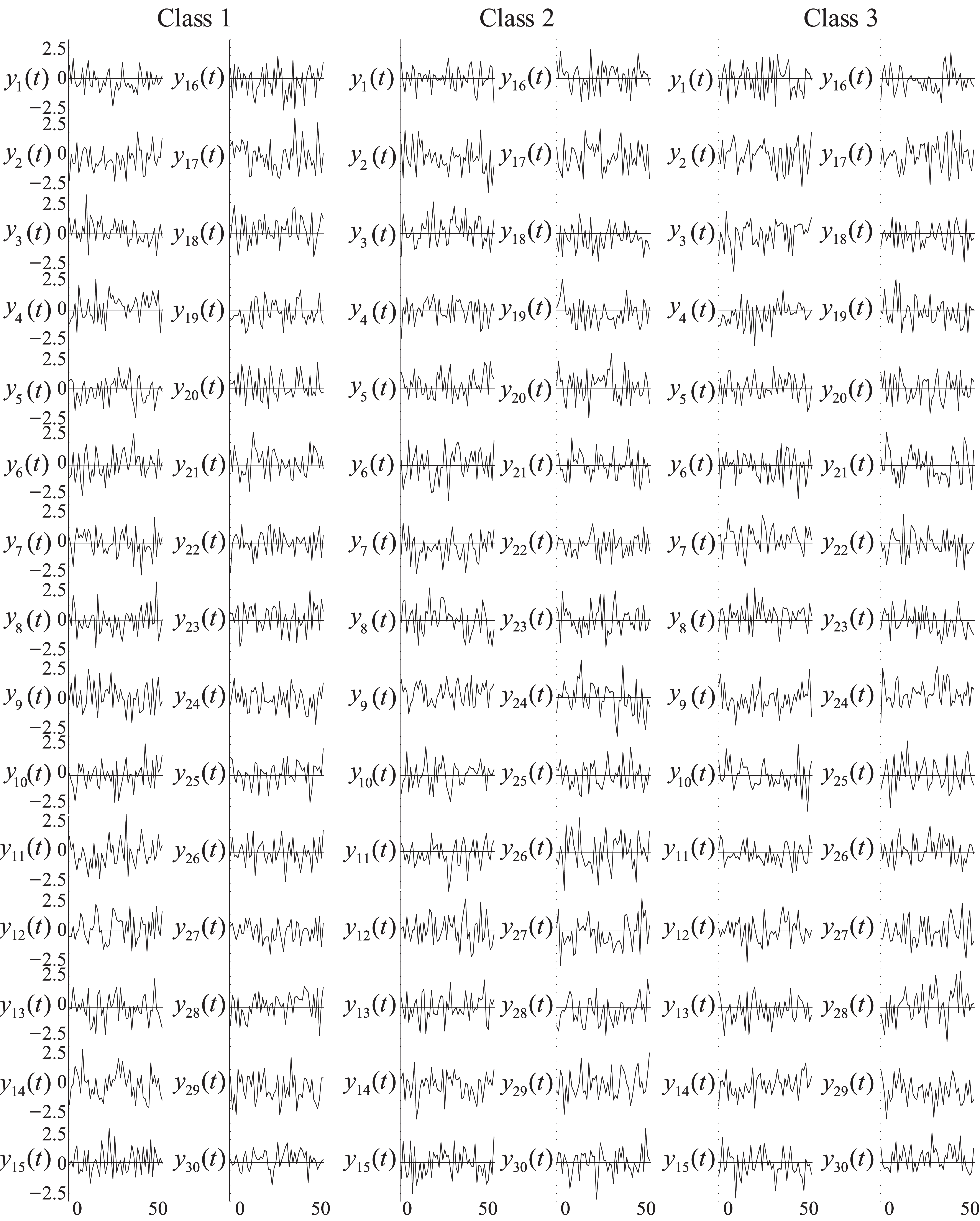}
	\caption{Example of ${\bm y}(t)$ used in the comparison with conventional dimensionality reduction techniques. 
			The data were prepared by combining ${\bm x}(t)$ (HMM-based signals) and white Gaussian noise ${\bm \eta} (t)$. 
			The noise ratio $a$ was $a=0.8$ in this figure.}
	\label{30dimNoise}
\end{figure}
Fig. \ref{PCAprob} shows an example of the problem used in comparison with PCA,  
where Fig. \ref{PCAprob} (a) is an example of time series $\bm{y}(t) \in {\Re}^2$, 
and Fig. \ref{PCAprob} (b) is a scattergram that indicates the distribution of population used in data generation.
Auxiliary axes $Z_1$ and $Z_2$ in Fig. \ref{PCAprob} (b) show the direction with the largest variance (the first principal component), 
and the direction orthogonal to $Z_1$ (the second principal component), respectively. 
The signals were generated based on the following equation so that $Z_1$ is an insignificant component for classification.  
\begin{equation}
y_i(t)
\!=\!\left\{ 
\begin{array}{l}
0.5\sin(2\pi t/100) + 0.5{\eta_i}(t) \hspace{4mm} {\rm for~class~1} \\
-0.5\sin(2\pi t/100) + 0.5{\eta_i}(t) \hspace{1.5mm}  {\rm for~class~2}
\end{array}
\right.,
\end{equation}
where $\bm{\eta}(t)$ is a Gaussian random number with a mean of 0 and a covariance matrix of [1 -1; -1 1], 
which is independently generated for each time and each class.
Reduced-dimensional signals were calculated using the TSDCN and PCA, 
and were compared with each other. 
For the reduced-dimensional signals of PCA, classification accuracy was then calculated using the R-LLGMN 
by changing the combination of 5 training samples and 100 test samples 10 times, 
and resetting the initial weight coefficients 10 times for each data set.
The classification accuracy of the TSDCN for the original problem was also calculated.

An example of the problem used in comparison with LDA is shown in Fig. \ref{LDAprob}, 
where Fig. \ref{LDAprob} (a) and (b) represent an example of a time series 
and a scattergram that indicates the distribution of population, respectively.
In Fig. \ref{LDAprob} (b), the data of class 1 are distributed in the first and third quadrants, 
and the data of class 2 are distributed in the second and fourth quadrants 
referring to the XOR problem that is a typical nonlinear classification problem. 
To be more precise, the data are uniform random numbers in the region where $(y_1>0 \land y_2>0 \land y_1+y_2<1) \lor (y_1<0 \land y_2<0 \land y_1+y_2>-1)$ for class 1, 
and $(y_1<0 \land y_2>0 \land -y_1+y_2<1) \lor (y_1>0 \land y_2<0 \land y_1-y_2<1)$ for class 2. 
In this problem, classification accuracy was also calculated in the same way as in the comparison with PCA.

Fig. \ref{30dimNoise} shows an example of the data ${\bm y} (t) \in {\Re}^{D}$ used to verify the influence 
of differences in dimensionality reduction methods on high-dimensional data classification. 
The data were produced by combining ${\bm x} (t)$ and white Gaussian noise ${\bm \eta} (t) \in {\Re}^{D}$ 
in the following way: 
\begin{equation}
\mbox{\boldmath $y$}(t) = (1-a){\bm x}(t) + a{\bm \eta}(t), 
\label{linear}
\end{equation}
where $a$ is a coefficient that determines noise ratio, and ${\bm x} (t)$ is an HMM-based signal used in Subsection A. 
The noise ${\bm \eta}(t)$ was added to verify the ability of each method for high-dimensional data 
that superposed an unnecessary component as the principal component.
The parameters used in this data generation are $C=3$, $D=30$, $T=50$, $a=0.8$.
The classification accuracy and training time of the TSDCN, PCA with the R-LLGMN, and LDA with the R-LLGMN are 
calculated in the same manner as in Subsection A. 

\subsubsection{Results}
\begin{figure}[t]
	\centering
	\includegraphics[width=1.0\hsize] {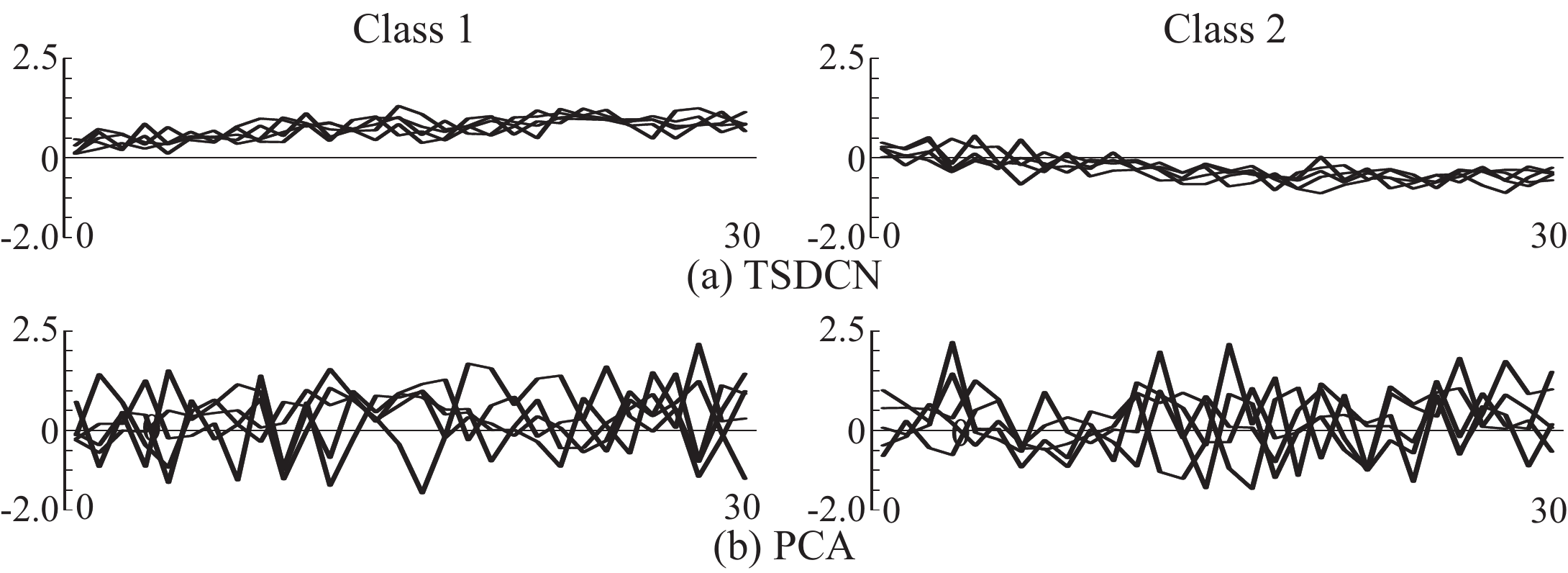}
	\caption{Reduced-dimensional signals of (a) the TSDCN and (b) PCA for the problem shown in Fig. \ref{PCAprob}.
			In (a), signals class 1 and class 2 seem to be separable. On the other hand, reduced-dimensional signals of PCA are
			similar to each other in (b).}
	\label{reducedPCA}
\end{figure}
\begin{figure}[t]
	\centering
	\includegraphics[width=1.0\hsize] {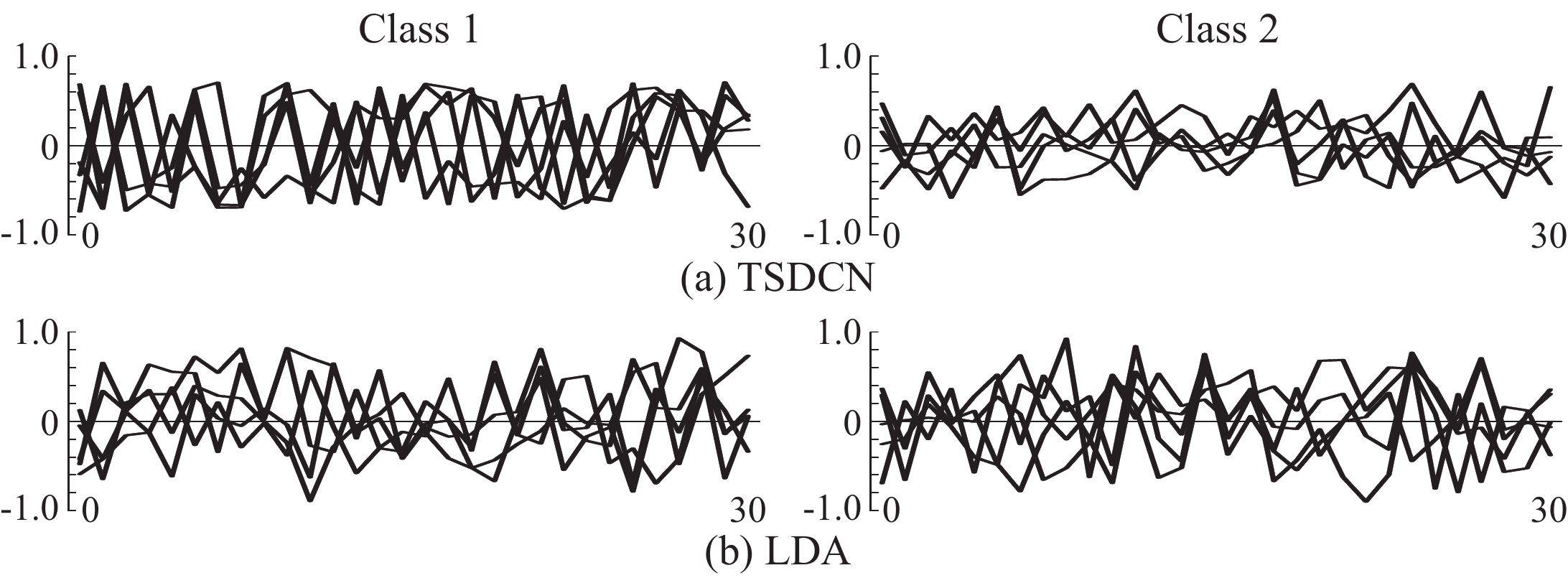}
	\caption{Reduced-dimensional signals of (a) the TSDCN and (b) LDA for the problem shown in Fig. \ref{LDAprob}.
			In (a), the amplitude of signals in class 1 seems greater than in class 2, 
			although visible separation of class 1 and class 2 was not observed. 
			In (b), however, the amplitude of signals in class 1 is almost the same as in class 2.}
	\label{reducedLDA}
\end{figure}

\begin{figure}[t]
	\centering
	\includegraphics[width=1.0\hsize] {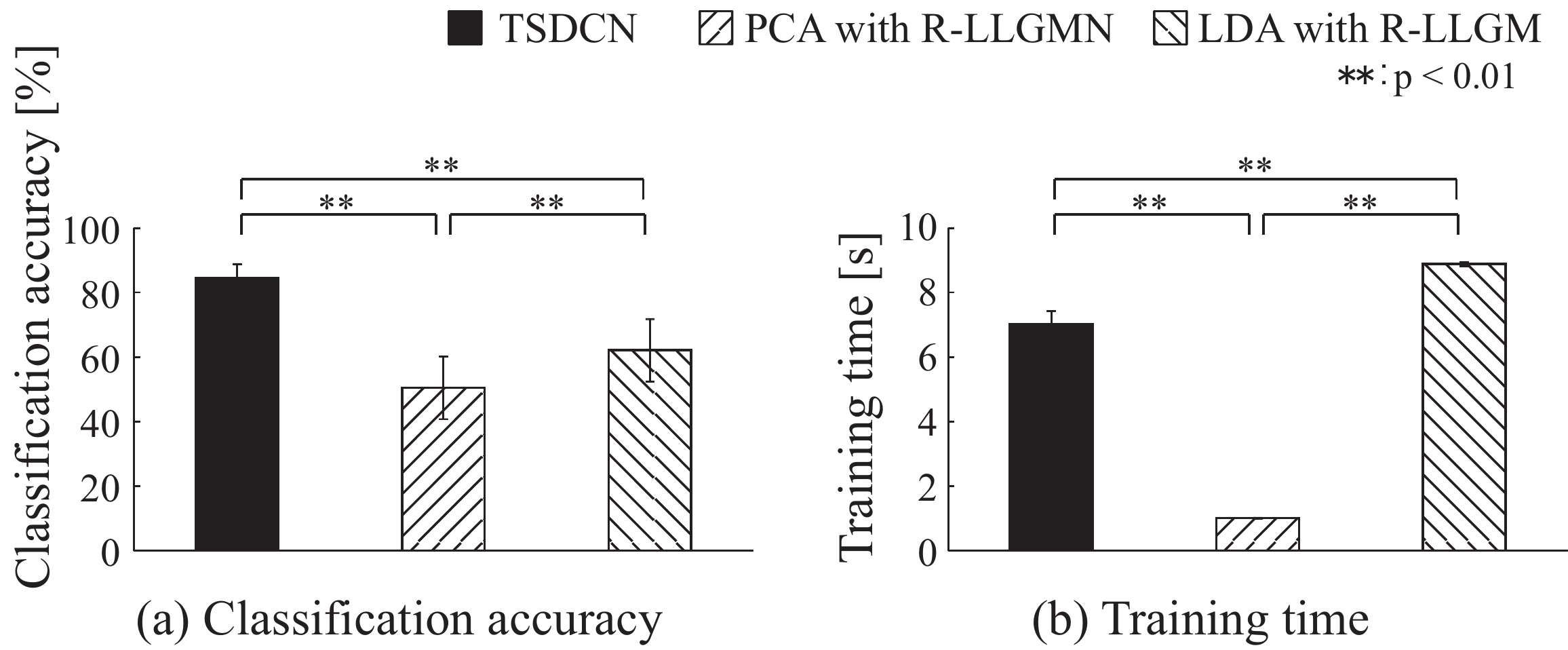}
	\caption{Average classification accuracy and training time for the comparison with conventional dimensionality reduction techniques}
	\label{AccuracyComp}
\end{figure}
Fig. \ref{reducedPCA} shows five examples of reduced-dimensional signals for the problem shown in Fig. \ref{PCAprob} 
using the TSDCN and PCA.
The classification accuracies of the TSDCN and PCA with the R-LLGMN were 
$100$ [\%] and $49.6 \pm 3.1$ [\%], respectively. 

Reduced-dimensional signals for Fig. \ref{LDAprob} using the TSDCN and LDA are shown in Fig. \ref{reducedLDA}. 
In Fig. \ref{reducedLDA} (a), the average standard deviations of signals in class 1 and class 2 were 
$0.49 \pm 0.03$ and $0.28 \pm 0.03$, respectively, and a significant difference was detected.
However, in Fig. \ref{reducedLDA} (b), there was no significant difference between 
class 1 ($0.40 \pm 0.04$) and class 2 ($0.40 \pm 0.04$).
The classification accuracies of the TSDCN and LDA with the R-LLGMN were 
$90.9 \pm 8.9$ [\%] and $66.1 \pm 9.6$ [\%], respectively, and a significant difference was confirmed. 

Fig. \ref{AccuracyComp} shows classification accuracy and training time for high-dimensional signals shown in Fig. \ref{30dimNoise}. 
The classification accuracies for the TSDCN, PCA with the R-LLGMN, and LDA with the R-LLGMN were 
$84.4 \pm 4.4$ [\%], $50.4 \pm 9.7$ [\%], and $62.1 \pm 9.7$ [\%], respectively.
The training times were $7.0 \pm 0.4$ [s] (the TSDCN), $1.0 \pm 0.0$ [s] (PCA with the R-LLGMN), 
and $8.8 \pm 0.1$ [s] (LDA with the R-LLGMN). 

\subsubsection{Discussion}
In Fig. \ref{reducedPCA} (a), differences between class 1 and class 2 can be observed: 
signals in class 1 showed positive values, and most of the signals in class 2 showed negative values. 
This is because the TSDCN obtained dimensionality reduction in the direction of the $Z_2$ axis (see Fig. \ref{PCAprob})
where data in class 1 and class 2 can be separated. 
In contrast, dimensionality reduction with PCA obtained a projection to the $Z_1$ axis 
where the variance of the data is merely the largest, 
hence the visible separation of class 1 and class 2 cannot be confirmed in Fig. \ref{reducedPCA} (b).
Thus, the TSDCN achieved a higher classification accuracy than PCA with R-LLGMN 
because of these differences in dimensionality reduction.

In Fig. \ref{reducedLDA} (a), the amplitude of signals in class 1 seems greater than in class 2, 
although the visible separation of class 1 and class 2 was not confirmed. 
However, in Fig. \ref{reducedLDA} (b), the amplitude of signals in class 1 is almost same as in class 2. 
Taking the significant difference in the standard deviation of reduced-dimensional signals, 
one possible explanation for the difference in classification accuracies is that 
the TSDCN contained the difference between class 1 and class 2 as amplitude information, 
whereas LDA could not be adapted to the nonlinear separation problem. 

These differences in dimensionality reduction affected the classification of high-dimensional signals 
(see Fig. \ref{AccuracyComp}). 
The TSDCN could acquire dimensionality reduction suitable for classification even if unnecessary components were superposed on the input data, 
and hence it achieved higher classification accuracy than other methods. 

These results suggested that the dimensionality reduction of the TSDCN has the capability
for the extraction of information necessary for classification from the input data that include unnecessary components, 
and the adaptation of a nonlinear separation problem to a certain extent.

\subsection{Comparison with our previous network}
\subsubsection{Method}
The data used in this comparison were the high-dimensional data that were the same as subsection C, i.e., data in Fig. \ref{30dimNoise}. 
Using 5 samples for training and 50 samples for testing, data sets were changed 10 times with resetting of the initial weight coefficients of the networks. 
The performance of TSDCA was then compared with that of the previous network in terms of accuracy, training time, and transition of log-likelihood. 

\subsubsection{Results}
\begin{figure}[t]
	\centering
	\includegraphics[width=1.0\hsize] {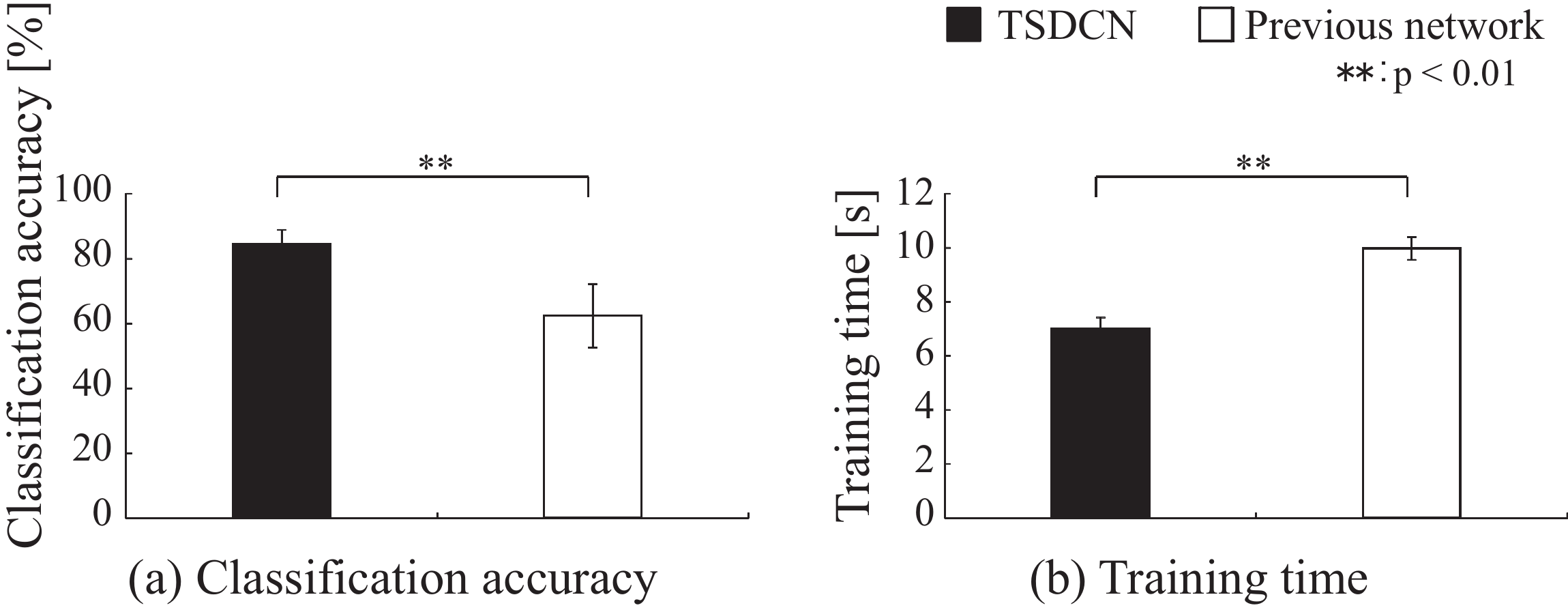}
	\caption{Average classification accuracy and training time for the comparison with our previous network}
	\label{AccuracyPrev}
\end{figure}

\begin{figure}[t]
	\centering
	\includegraphics[width=1.0\hsize] {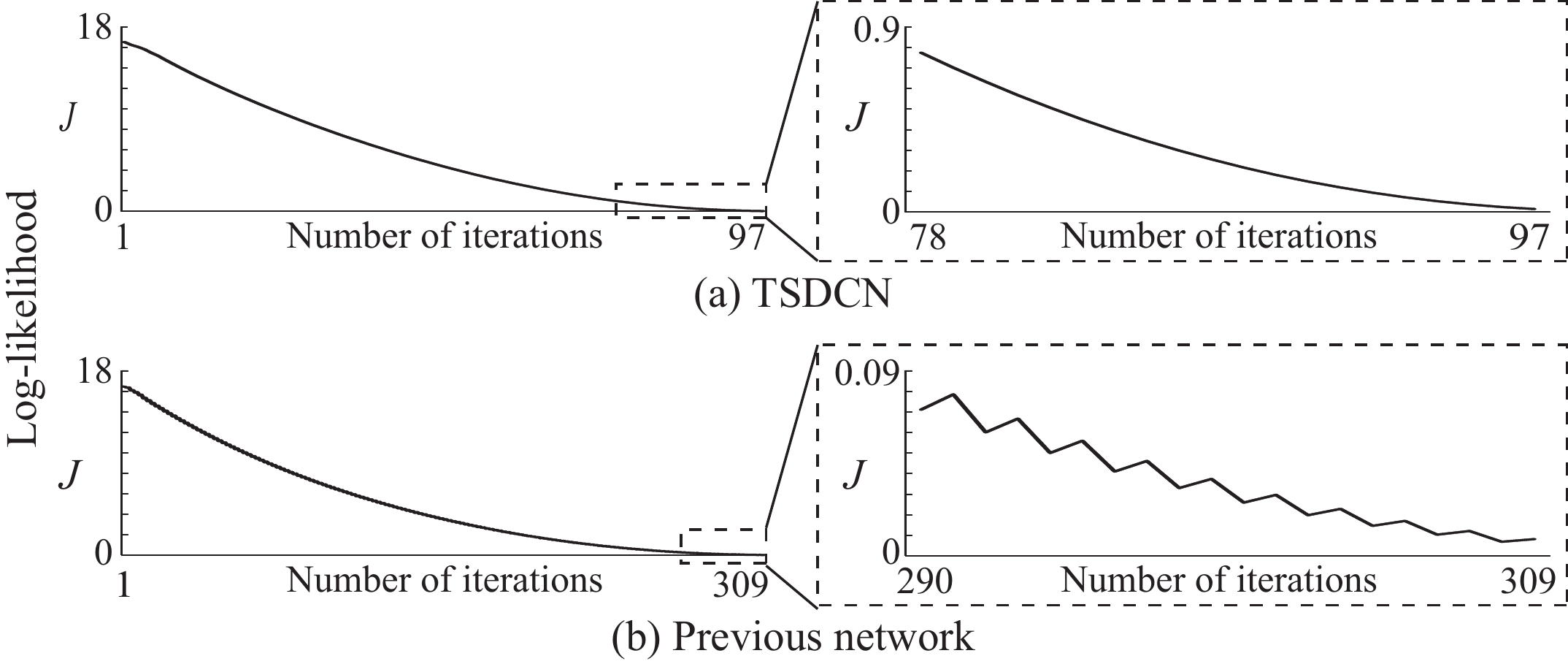}
	\caption{Examples of the transition of log-likelihood}
	\label{loglikelihood}
\end{figure}
Fig. \ref{AccuracyPrev} shows the average classification accuracy and training time for each method. 
The classification accuracies of the TSDCN and the previous network were $84.4 \pm 4.4$ [\%] and $62.4 \pm 9.8$ [\%], 
and a significant difference was confirmed ($p < 0.01$). 
The difference in the training time between the TSDCN ($7.0 \pm 0.4$ [s]) and the previous network ($10.0 \pm 0.4$ [s]) was also significant. 
Examples of the transition of log-likelihood $J$ to the number of training iterations are given in Fig. \ref{loglikelihood}. 
Fig. \ref{loglikelihood} (a) and (b) are $J$ of the TSDCN and the previous network, respectively. 
The left panels show all iterations until the end of learning, and the right ones show the magnifications of the last 20 iterations.

\subsubsection{Discussion}
From the significant difference in Fig. \ref{AccuracyPrev}, 
the classification accuracy of the TSDCN was improved over the previous network, 
although this is nothing more than a single example. 
The result that should be emphasized is the improvement of convergency. 
In Fig. \ref{loglikelihood}, the log-likelihood $J$ of the TSDCN is smoothly decreasing, 
whereas that of the previous network is jaggedly decreasing. 
The TSDCN can reduce the log-likelihood monotonically because of the Lagrange multiplier-based learning. 
The previous network, however, alternately calculates the weight modulations based on the gradient descent that decreases $J$ 
and the Gram-Schmidt process for orthogonalization that increases $J$, 
hence the curve of $J$ becomes jagged. 
If the increase exceeds the decrease, the learning cannot converge. 
The improvement of the convergency can also be seen in the difference in the training time.

\section{EEG classification}
To evaluate the applicability of the TSDCN for real biological data, 
a classification experiment was conducted using two EEG datasets. 
One was a 64-channel ($D=64$) EEG dataset downloaded from the UCI Machine Learning Repository \cite{Bache1999uci} (dataset 1). 
The EEG signals were recorded from a healthy male who was exposed two classes of stimuli 
($C=2$: (a) a single stimulus, (b) two stimuli) for 1 second, 10 times each class. 
The stimuli were pictures of objects chosen from the Snodgrass and Vanderwart picture set \cite{snodgrass1980standardized}. 
3 samples were then used as a training set, and 7 samples were treated as a test set in the classification experiment. 
The other was a 19-channel ($D=19$) EEG dataset downloaded from Project BCI - EEG motor activity data set \cite{midhat2008project} (dataset 2).
The EEG was recorded from a healthy 21-year-old male who performed actual random movements of left and right hand ($C=2$) for about 128.6 seconds each. 
In the classification experiment, with 0.1 seconds of data as a sample for each class, 50 samples were used as a training set and the remaining 1,236 were treated as a test set.
With respect to both datasets, average classification rate and training time were then calculated 
by changing the combination of training/test data sets randomly 10 times 
and resetting the initial weight coefficients 10 times for each combination, 
and compared with those of an HMM, the R-LLGMN, PCA with the R-LLGMN, LDA with the R-LLGMN, the Elman network, and classification based on common spatial patterns (CSP) \cite{muller1999designing}. 
CSP is a feature extraction algorithm for EEG that was originally introduced by Koles {\it et al.} \cite{koles1990spatial}. 
We implemented a classification method according to the method of M{\"u}ller-Gerking {\it et al.}\cite{muller1999designing}, 
which is the combination of a CSP-based feature extraction and a linear Bayesian classifier.
The parameters of the TSDCN were $K_c=2$, $M_{c,k} = 2$, $D'=1$. 
Other experimental conditions were same as the experiments conducted in Section IV. 

\begin{figure}[t]
	\centering
	\includegraphics[width=1.0\hsize] {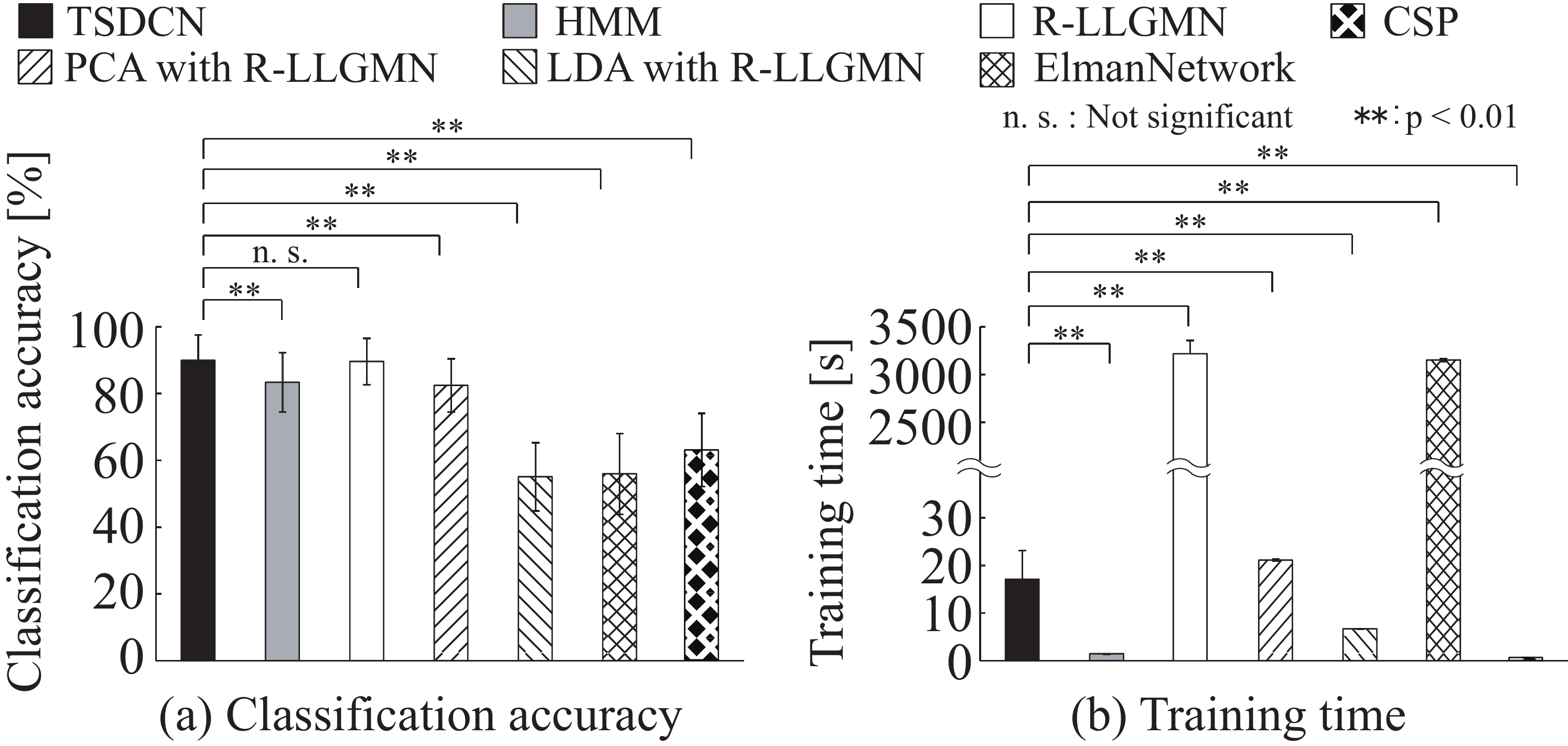}
	\caption{Average classification accuracy and training time of each method for the EEG dataset 1}
	\label{AccuracyEEG}
\end{figure}

\begin{figure}[t]
	\centering
	\includegraphics[width=1.0\hsize] {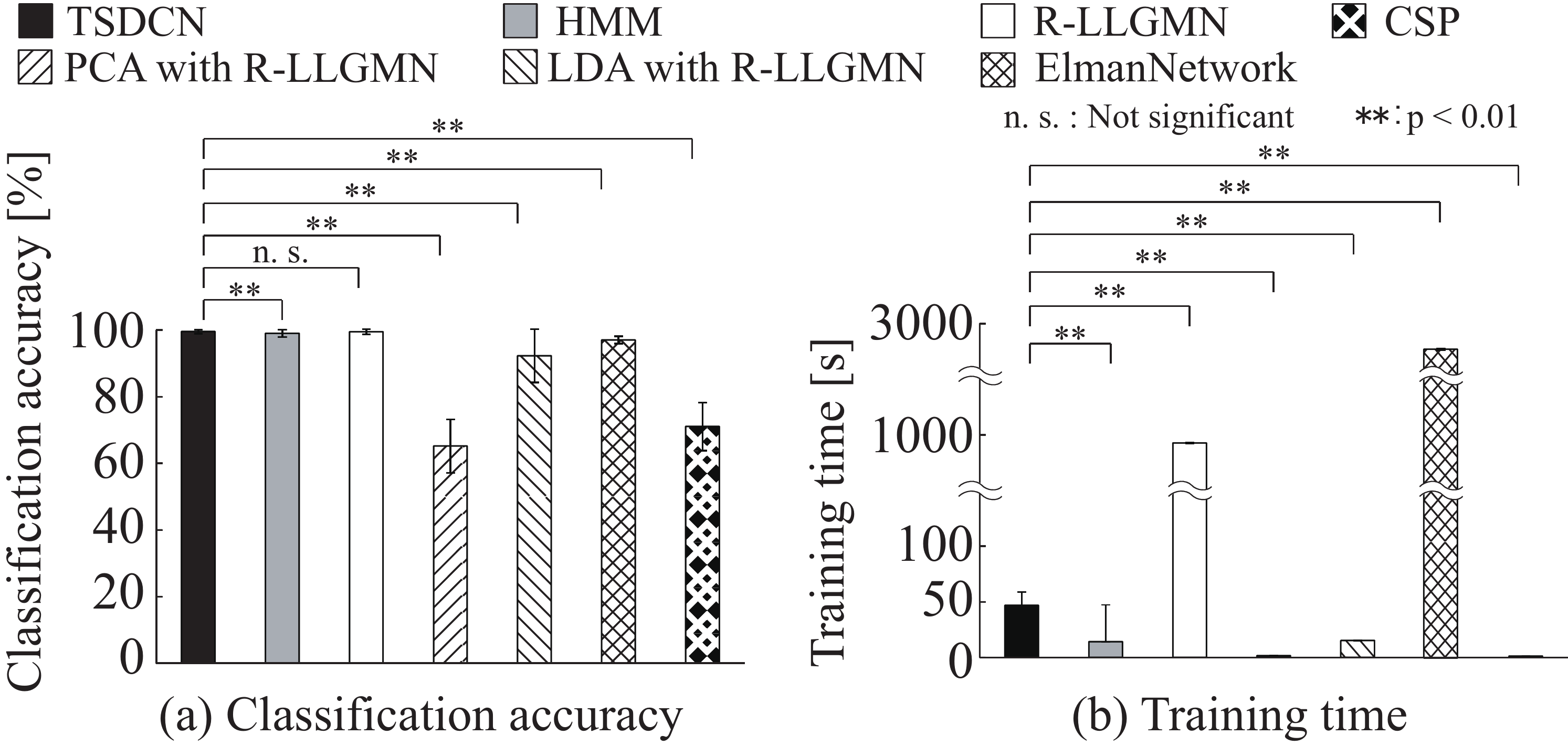}
	\caption{Average classification accuracy and training time of each method for the EEG dataset 2}
	\label{AccuracyBCI}
\end{figure}

Fig. \ref{AccuracyEEG} shows the average classification accuracy and training time for dataset 1. 
The results of significant differences between the TSDCN and the other methods are also shown in Fig. \ref{AccuracyEEG}. 
The classification accuracies of the TSDCN, the HMM, the R-LLGMN, PCA with the R-LLGMN, LDA with the R-LLGMN, the Elman network, and CSP 
were $89.8 \pm 7.6$ [\%], $83.4 \pm 8.9$ [\%], $89.6 \pm 7.0$ [\%], $82.4 \pm 8.0$ [\%], $55.0 \pm 10.2$ [\%], $55.9 \pm 12.1$ [\%], and $62.9 \pm 11.0$ [\%], respectively. 
The training times were reduced in the order of the R-LLGMN ($3215.0 \pm 140.5$ [s]), the Elman network ($3150.5 \pm 15.1$ [s]), 
PCA with the R-LLGMN ($21.1 \pm 0.2$ [s]), the TSDCN ($17.1 \pm 6.0$ [s]), LDA with the R-LLGMN ($6.7 \pm 0.0$ [s]), the HMM ($1.4 \pm 0.0$ [s]), and CSP ($0.7 \pm 0.0$ [s]).

The classification accuracy and training time for dataset 2 are shown in Fig. \ref{AccuracyBCI}. 
The classification accuracies of the TSDCN, the HMM, the R-LLGMN, PCA with the R-LLGMN, LDA with the R-LLGMN, the Elman network, and CSP 
were $99.5 \pm 0.6$ [\%], $99.0 \pm 1.1$ [\%], $99.5 \pm 0.8$ [\%], $65.2 \pm 1.5$ [\%], $92.2 \pm 15.2$ [\%], $97.0 \pm 1.1$ [\%], and $71.0 \pm 1.1$ [\%], respectively. 
The training times of the TSDCN, the HMM, the R-LLGMN, PCA with the R-LLGMN, LDA with the R-LLGMN, the Elman network, and CSP 
were $46.8 \pm 12.0$ [s], $14.2 \pm 32.8$ [s], $963.5 \pm 1.1$ [s], $1.5 \pm 0.0$ [s], $15.2 \pm 0.1$ [s], $2780.0 \pm 8.1$ [s], and $1.1 \pm 0.0$ [s], respectively. 

These results revealed that the TSDCN 
can reduce training time drastically compared with conventional NNs such as the R-LLGMN and the Elman network, 
and can compress and classify high-dimensional time-series EEG signals with a relatively high accuracy.

\section{Conclusion}
This paper proposed a novel recurrent probabilistic neural network called the TSDCN. 
The TSDCN is developed on the basis of TSDCA that consists of several orthogonal transformation matrices 
and HMMs that include GMMs for probabilistic densities, 
thereby allowing dimensionality reduction of input data and calculation of posterior probabilities for each class. 
The backpropagation through time-based learning algorithm improved by integrating the Lagrange multiplier method
enables the network to obtain the parameters of dimensionality reduction and classification simultaneously. 

In the simulation experiments conducted in this paper, 
we revealed the following facts about TSDCN: 
\begin{itemize}
\item The TSDCN can classify high-dimensional time-series data accurately, even for a high multi-class problem, if the input data obey HMMs.
\item The increase of the amount of calculation based on the increase in input dimensionality can be suppressed to linear 
if the number of reduced dimensions is sufficiently small.
\item The TSDCN can reduce training time while maintaining classification accuracy equivalent to conventional NNs.
\item The dimensionality reduction of the TSDCN has the ability to extract the information necessary for classification 
and adapt the nonlinear separation problem to a certain extent.
\item Because of the Lagrange multiplier-based learning, the convergency of the learning is improved over our previous network.
\end{itemize}

The applicability of the TSDCN for real biological data is also suggested in the EEG data classification experiments using 2 datasets. 
The results of the experiment showed that the TSDCN demonstrated a high classification performance ($89.8 \pm 7.6$ [\%], $99.5 \pm 0.6$ [\%]) 
and relatively fast learning ($17.1 \pm 6.0$ [s], $46.8 \pm 12.0$ [s]). 

In future research, the authors would like to adapt this approach to other applications such as video recognition and brain--computer interfaces. 
We also plan to introduce the kernel trick to the model structure to improve the dimensionality reduction ability 
for nonlinear separation problems.

\appendix[Gradient calculation]
The backpropagation-through-time (BPTT) algorithm \cite{werbos1990backpropagation} is used for the gradient calculation. 
Using the chain rule, each element of (\ref{w1}) and (\ref{w2}) is expanded in the following way:
\begin{eqnarray}
\lefteqn{\frac{\partial J_n}{\partial w_{i,j}^{(c,k,m)}}}\nonumber\\
&=&-\sum^{T-1}_{t=0}\sum^{C}_{c'=1}\sum^{K_{c'}}_{k''=1}\Delta^{c'}_{k''}(t)
	\frac{\partial {}^{(6)}O^{c'}_{k''}(T-t)}{\partial 	{}^{(6)}I^{c}_{k}(T-t)}\nonumber \\
&\times&\sum^{K_c}_{k'=1}\frac{\partial {}^{(6)}I^{c}_{k}(T-t)}{\partial {}^{(5)}O^{c}_{k',k}(T-t)}
	\frac{\partial {}^{(5)}O^{c}_{k',k}(T-t)}{\partial {}^{(5)}I^{c}_{k',k}(T-t)}\nonumber\\
&\times&\frac{\partial {}^{(5)}I^{c}_{k',k}(T-t)}{\partial {}^{(4)}O^{c}_{k',k,m}(T-t)}
	\frac{\partial {}^{(4)}O^{c}_{k',k,m}(T-t)}{\partial {}^{(4)}I^{c}_{k',k,m}(T-t)}\nonumber\\
&\times&\sum^{H}_{h=1}\frac{\partial {}^{(4)}I^{c}_{k',k,m}(T-t)}{\partial {}^{(3)}O^{c,k,m}_h(T-t)}
	\frac{\partial {}^{(3)}O^{h}_{c,k,m}(T-t)}{\partial {}^{(2)}I^{c,k,m}_j(T-t)}  \nonumber \\
&\times&\frac{\partial {}^{(2)}I^{j}_{c,k,m}(T-t)}{\partial w_{i,j}^{(c,k,m)}(T-t)}\nonumber\\
&=&-\sum^{T-1}_{t=0}\sum^{C}_{c'=1}\sum^{K_{c'}}_{k''=1}\Delta^{c'}_{k''}(t)\nonumber\\
& &\times(\delta_{(c',k''),(c,k)}-{}^{(6)}O^{c'}_{k''}(T-t))
	\frac{{}^{(6)}O^{c'}_{k''}(T-t)}{{}^{(6)}I^{c'}_{k''}(T-t)}\nonumber\\
& &\times\sum^{K_c}_{k'=1}{}^{(6)}O^{c}_{k'}(T-t-1){}^{(4)}O^{c}_{k',k,m}(T-t)\nonumber\\
& &\times\sum^{H}_{h=1}{}^{(2)}w_h^{(c,k',k,m)}\frac{\partial {}^{(3)}O^{h}_{c,k,m}(T-t)}{\partial {}^{(2)}I^{c,k,m}_j(T-t)}\nonumber\\
& &\times{}^{(1)}O_i(T-t),
\end{eqnarray}
\begin{eqnarray}
\lefteqn{\frac{\partial J_n}{\partial {w'}_h^{(c,k',k,m)}}}\nonumber\\
&=&-\sum^{T-1}_{t=0}\sum^{C}_{c'=1}\sum^{K_{c'}}_{k''=1}\Delta^{c'}_{k''}(t)
	\frac{\partial {}^{(6)}O^{c'}_{k''}(T-t)}{\partial 	{}^{(6)}I^{c}_{k}(T-t)}\nonumber \\
&\times&\frac{\partial {}^{(6)}I^{c}_{k}(T-t)}{\partial {}^{(5)}O^{c}_{k',k}(T-t)}
	\frac{\partial {}^{(5)}O^{c}_{k',k}(T-t)}{\partial {}^{(5)}I^{c}_{k',k}(T-t)}\nonumber\\
&\times&\frac{\partial {}^{(5)}I^{c}_{k',k}(T-t)}{\partial {}^{(4)}O^{c}_{k',k,m}(T-t)}
	\frac{\partial {}^{(4)}O^{c}_{k',k,m}(T-t)}{\partial {}^{(4)}I^{c}_{k',k,m}(T-t)}\nonumber\\
&\times&\frac{\partial {}^{(4)}I^{c}_{k',k,m}(T-t)}{\partial {w'}_h^{(c,k',k,m)}} \nonumber \\
&=&-\sum^{T-1}_{t=0}\sum^{C}_{c'=1}\sum^{K_{c'}}_{k''=1}\Delta^{c'}_{k''}(t)\nonumber\\
& &\times(\delta_{(c',k''),(c,k)}-{}^{(6)}O^{c'}_{k''}(T-t))
	\frac{{}^{(6)}O^{c'}_{k''}(T-t)}{{}^{(6)}I^{c'}_{k''}(T-t)}\nonumber\\
& &\times{}^{(6)}O^{c}_{k'}(T-t-1){}^{(4)}O^{c}_{k',k,m}(T-t)\nonumber\\
& &\times{}^{(2)}O^h_{c,k,m}(T-t),
\end{eqnarray}
where $\Delta^{c'}_{k''}(t)$ is defined as the partial differentiation of $J_n$ to ${}^{(6)}O^{c'}_{k''}(T-t)$.
\begin{equation}
	\Delta^{c'}_{k''}(t) = \frac{\partial J_n}{\partial {}^{(6)}O^{c'}_{k''}(T-t)}.
\label{S3D3}
\end{equation}
(\ref{S3D3}) can be derived as follows:
\begin{eqnarray}
{\Delta^{c'}_{k''}(0)}
&=& \frac{\partial (Q_{c'}\log{{}^{(7)}O^{c'}(T)})}{\partial {}^{(7)}O^{c'}(T)}
	\frac{\partial {}^{(7)}O^{c'}(T)}{\partial {}^{(6)}O^{c'}_{k''}(T)}\ \ \ \ \ \ \ \ \ \  \nonumber \\
&=& \frac{Q_{c'}}{{}^{(7)}O^{c'}(T)},
\label{S3D4}
\end{eqnarray}
\begin{eqnarray}
\Delta^{c'}_{k''}(t+1)
&\!\!\!\!\!\!=\!\!\!\!\!\!& \sum_{c''=1}^C\sum_{k'''=1}^{K_{c''}}\Delta^{c''}_{k'''}(t)\sum_{k''''=1}^{K_{c''}}
	\frac{\partial {}^{(6)}O^{c''}_{k'''}(T-t)}{\partial {}^{(6)}I_{k''''}^{c'}(T-t)} \nonumber \\
	&\!\!\! \!\!\!& \times\frac{\partial {}^{(6)}I_{k''''}^{c'}(T\!-\!t)}{\partial {}^{(5)}O^{c'}_{k'',k''''}(T\!-\!t)}
	\frac {\partial {}^{(5)}O^{c'}_{k'',k''''}(T\!-\!t)}{\partial {}^{(6)}O^{c'}_{k''}(T\!-\!(t\!+\!1))} \nonumber \\
&\!\!\!\!\!\!=\!\!\!\!\!\!& \sum_{c''=1}^C\sum_{k'''=1}^{K_{c''}}\Delta^{c''}_{k'''}(t) \nonumber \\
	&\!\!\! \!\!\!&\times\!\sum_{k''''=1}^{K_{c''}}\!\!(\delta_{(c'',k'''),(c',k'''')}\!-\!{}^{(6)}O^{c''}_{k'''}(T\!-\!t)) \nonumber \\
	&\!\!\! \!\!\!&\times\frac{{}^{(6)}O^{c''}_{k'''}(T-t)}{{}^{(6)}I^{c''}_{k'''}(T-t)}{}^{(5)}I^{c'}_{k'',k''''}(T-t).
\label{S3D5}
\end{eqnarray}
Moreover, the partial differentiation of ${}^{(3)}O^{c,k,m}_h(T-t)$ to ${}^{(2)}I^{j}_{c,k,m}(T-t)$ can be calculated as
\begin{eqnarray}
\lefteqn{\frac{\partial {}^{(3)}O^{c,k,m}_h(T-t)}{\partial {}^{(2)}I^{j}_{c,k,m}(T-t)}}\nonumber \\
&=&\left\{ 
\begin{array}{l}
2{}^{(2)}I^{j}_{c,k,m}(T-t)
	\hspace{8mm}(j=s=l) \\
{}^{(2)}I^{l}_{c,k,m}(T-t)
	\hspace{10mm}(j=s, j\neq l)\\
{}^{(2)}I^{s}_{c,k,m}(T-t)
	\hspace{10mm}(j\neq s, j=l)\\
0\hspace{31.5mm}({\rm otherwise})
\end{array}
\right.,
\end{eqnarray}
where $h,s,l$ are integers, which satisfy (\ref{int_eq}).
\begin{equation}
\label{int_eq}
h= -\frac{1}{2}s^2 + (D'+\frac{1}{2})s+l-D'+1
\hspace{5mm}(1\leq s \leq l \leq D').
\end{equation}

The partial differentiation ${\partial h^{(c,k,m)}_i }/{\partial {\bf W}^{(c,k,m)}}$ in (\ref{parL1}), 
in addition, is calculated as 
\begin{eqnarray}
\frac{\partial h^{(c,k,m)}_i }{\partial {\bf W}^{(c,k,m)}}
\!=\! \left[\!\!\!
\begin{array}{cccc}
\bm{0}, & (\frac{\partial h^{(c,k,m)}_i }{{\bm{v}_1^{(c,k,m)}}})^{\rm T}, & \cdots, & (\frac{\partial h^{(c,k,m)}_i }{{\bm{v}_{D'}^{(c,k,m)}}})^{\rm T} \\
\end{array} 
\!\!\!\right]^{\rm T}\!\!,\!
\end{eqnarray}
where 
\begin{eqnarray}
\frac{\partial h^{(c,k,m)}_i }{{\bm{v}_{j'}^{(c,k,m)}}}
 &\!\!\!=\!\!\!& \frac{\partial ({\bm{v}^{(c,k,m)}_j}^{\rm T}{\bm{v}^{(c,k,m)}_l} - \delta_{j,l}) }{{\bm{v}_{j'}^{(c,k,m)}}} \nonumber \\
 &\!\!\!=\!\!\!& \left\{ 
\begin{array}{l}
2{\bm{v}_{j'}^{(c,k,m)}}
	\hspace{17mm}(j'=j=l) \\
{\bm{v}_{l}^{(c,k,m)}}(T-t)
	\hspace{8mm}(j'=j, j' \neq l)\\
{\bm{v}_{j}^{(c,k,m)}}(T-t)
	\hspace{8mm}(j' \neq j, j'=l)\\
0\hspace{28.5mm}({\rm otherwise})
\end{array}
\right..
\end{eqnarray}

}

\ifCLASSOPTIONcaptionsoff
  \newpage
\fi

\end{document}